\documentclass{article}

    \PassOptionsToPackage{numbers, compress}{natbib}
 \usepackage[preprint]{neurips_2026}

\usepackage[utf8]{inputenc} 
\usepackage[T1]{fontenc}    
\usepackage{hyperref}       
\usepackage{url}            
\usepackage{graphicx}       
\usepackage{booktabs}       
\usepackage{multirow}       
\usepackage{amsfonts}       
\usepackage{amssymb}        
\usepackage{nicefrac}       
\usepackage{microtype}      
\usepackage{xcolor}         
\usepackage{makecell}
\usepackage{subcaption}
\usepackage{amsmath}
\usepackage{enumitem}
\usepackage{listings}
\lstdefinelanguage{json}{
    basicstyle=\ttfamily\small,
    morestring=[b]",
    morecomment=[l]{//},
}

\title{TrajPrism: A Multi-Task Benchmark for Language-Grounded Urban Trajectory Understanding}

\author{%
  Lihuan Li$^{1\dagger}$\thanks{Equal contribution.\quad $^\dagger$Corresponding author: lihuan.li@unsw.edu.au.} \quad
  Wilson Wongso$^{1\ast}$ \quad
  Baiyu Chen$^{1}$ \quad
  Hao Xue$^{1,2}$ \quad
  Ruiyi Yang$^{1}$ \\
  {\bfseries Yifan Duan$^{1}$ \quad
  Xiachong Lin$^{1}$ \quad
  Yang Song$^{1}$ \quad
  Flora Salim$^{1}$} \\[6pt]
  $^1$UNSW Sydney\quad
  $^2$HKUST (GZ) \\[4pt]
}

\begin{document}

\maketitle

\begin{abstract}
  Urban mobility is naturally expressed both as trajectories in space and as natural-language descriptions of travel intent, constraints, and preferences.
  However, prior work rarely evaluates these two modalities \emph{together} on the \emph{same} real-world trajectories: trajectory modeling often stays geometry--centric, 
  while language--centric mobility benchmarks frequently target route planning and tool use rather than fine-grained, verifiable alignment between text and the underlying route.
  We introduce \textbf{TrajPrism}, a multi-task benchmark for language-trajectory alignment that unifies 
  (i) instruction-conditioned trajectory generation, 
  (ii) language-driven semantic trajectory retrieval, 
  and (iii) trajectory captioning, together with an evaluation protocol that measures trajectory fidelity, retrieval quality, and language groundedness.
  We construct TrajPrism by pairing real urban trajectories with judge-filtered language annotations generated under a four-dimensional travel-intent taxonomy.
  The benchmark contains 300K selected trajectories across Porto, San Francisco, and Beijing, yielding 2.1M task instances from three instruction variants, three retrieval queries, and one caption per trajectory.
  We further develop proof-of-concept models for each task: \textbf{TrajAnchor} for instruction-conditioned trajectory generation, \textbf{TrajFuse} for semantic trajectory retrieval, and \textbf{TrajRap} for trajectory captioning. 
  These models instantiate the proposed tasks and show that geometry-only trajectory baselines leave a large gap on our protocol, 
  especially where language is part of the input--output interface.
  We release TrajPrism with code and a reproducible annotation pipeline that is designed to be portable across cities, 
  given compatible trajectory inputs and map resources.
  Code available at \url{https://anonymous.4open.science/r/TrajPrism-05D6/}.
\end{abstract}

\section{Introduction}

Urban trajectories are one of the richest sources of human behavioral data available at scale.
Understanding them unlocks a broad range of applications: 
route planning~\cite{cao2025holistic}, mobility pattern analysis~\cite{wongso2025massive,wang2024large}, 
urban trajectory retrieval and similarity search~\cite{li2024t,chang2023contrastive,zhou2025blurred}, 
and location-based services~\cite{wongso2025genup}. \looseness -1

Large language models have accelerated interest in language interfaces for mobility~\cite{li2024urbangpt,han2025large}, 
pushing trajectory understanding beyond purely geometric prediction toward settings where models must interpret and produce natural language tied to observed trips.
To study this paradigm, recent benchmarks fragment this space along different axes.
Semantic Routing~\cite{zhao2024semantic} studies natural-language criteria paired with route-oriented generation. 
MobilityBench~\cite{song2026mobilitybench} evaluates LLM agents on realistic route-planning and information-access episodes expressed as user queries.
Both are valuable, but they do not jointly operationalize \textbf{language--trajectory alignment} on the same collection of real observed trajectories:
Semantic Routing centers on a generation-focused setting with synthetic query distributions,
while MobilityBench focuses on tool-mediated agent planning rather than language-grounded trajectory understanding.
What remains missing is a benchmark that 
(i) binds text to concrete trajectories, 
(ii) covers both language-to-trajectory and trajectory-to-language interfaces, 
and (iii) supports unified evaluation of trajectory fidelity, retrieval, and language groundedness,
going beyond the single-interface settings offered by existing work. \looseness -1

\begin{figure}[t]
    \centering
    \includegraphics[width=\linewidth]{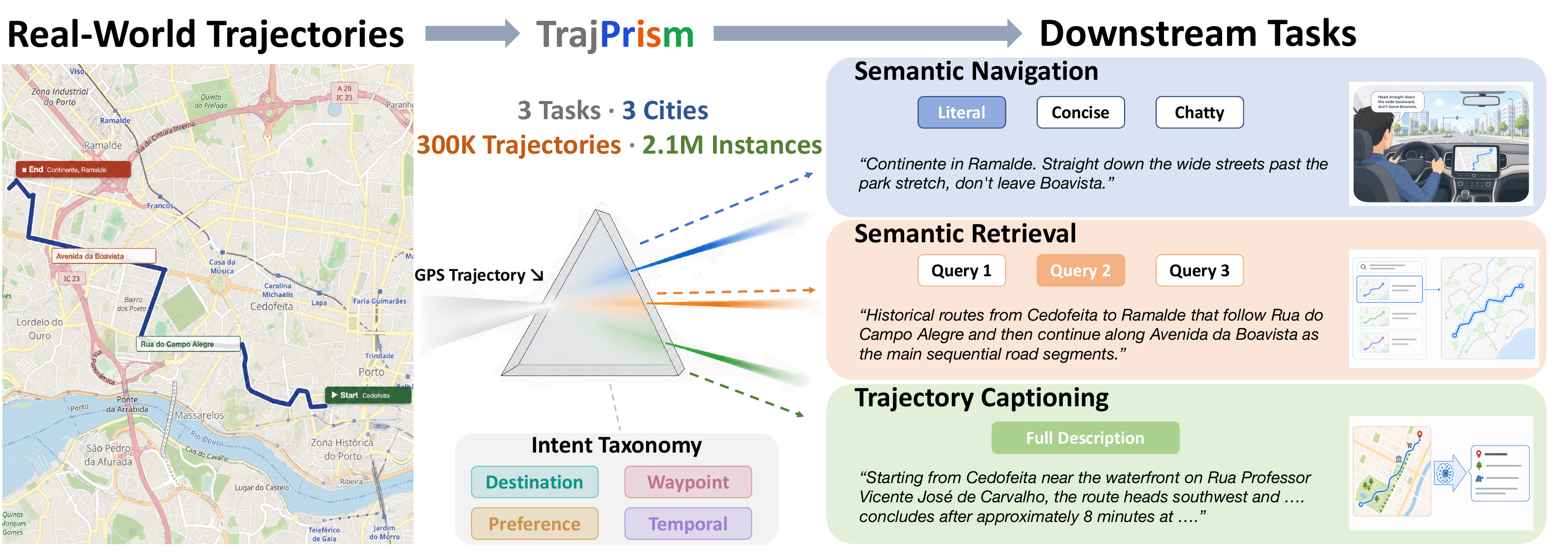}
    \caption{TrajPrism refracts real urban trajectories through a
    four-dimensional intent taxonomy into diverse, grounded language annotations.
    The benchmark supports three complementary tasks: navigation instruction
    following, trajectory retrieval, and trajectory captioning, across
    300K trajectories and three cities.}
    \label{fig:intro}
    \vspace{-16pt}
\end{figure}

\begin{table}[t]
    \centering
    \caption{Comparison with closely related benchmarks. TrajPrism is the only one built from real observed trajectories with bidirectional language--trajectory tasks.}
    \label{tab:related_benchmarks}
    \small
    \setlength{\tabcolsep}{3pt}
    \renewcommand{\arraystretch}{1.08}
    \begin{tabular}{@{}p{0.21\linewidth}p{0.19\linewidth}p{0.19\linewidth}p{0.30\linewidth}@{}}
    \toprule
    \textbf{Property} & \textbf{Sem.\ Routing}~\cite{zhao2024semantic} & \textbf{MobilityBench}~\cite{song2026mobilitybench} & \textbf{TrajPrism (ours)} \\
    \midrule
    Data source & Synthetic queries & Real user queries & Real GPS trajectories \\
    City diversity & U.S. cities & Global long-tail & Europe / U.S. / Asia \\
    Scale & 1M queries & 100K queries & \textbf{300K trajs. / 2.1M instances} \\
    Lang. grounding & Text \(\rightarrow\) route & Text \(\rightarrow\) API calls & \textbf{Text \(\leftrightarrow\) GPS trajectory} \\
    Tasks & Route generation & Agent planning & \textbf{Gen. + retrieval + captioning} \\
    Public pipeline & \(\times\) & \(\times\) & \(\checkmark\) \\
    \bottomrule
    \end{tabular}
    \vspace{-20pt}
\end{table}

We introduce TrajPrism to fill this gap.
TrajPrism targets language--trajectory alignment through three complementary tasks:
\textit{instruction-conditioned trajectory generation} that asks a model to predict a trajectory from a natural-language navigation instruction,
\textit{language-driven semantic trajectory retrieval} that challenges a model to identify the matching trajectory from a candidate pool given a textual query,
and \textit{trajectory captioning} that requires a model to produce a factual description of an observed trajectory.
Together, these tasks probe whether models can align unconstrained navigation language with urban routes.
This is analogous in spirit to vision--language benchmarks~\cite{lin2014microsoft,plummer2015flickr30k} that unify generation, retrieval, and description, but instantiated on urban mobility.
Figure~\ref{fig:intro} illustrates the core idea: real urban trajectories are refracted through a four-dimensional intent taxonomy into diverse, grounded language annotations that feed the three tasks.
Table~\ref{tab:related_benchmarks} contrasts TrajPrism with the most closely related benchmarks.
Whereas Semantic Routing and MobilityBench each cover a single evaluation interface (route generation or agent planning, respectively), 
TrajPrism is the only benchmark built from real GPS trajectories that supports bidirectional language--trajectory tasks and provides a public generation pipeline.

Our main contributions are as follows:
\begin{itemize}[leftmargin=*,itemsep=2pt,topsep=2pt]

    \item We introduce \textbf{TrajPrism}, the first large-scale, multi-task,
    language-grounded benchmark for urban trajectory understanding.
    TrajPrism pairs 300K real-world trajectories across three cities
    with 2.1M task instances (each trajectory yields seven instances:
    three instruction variants, three retrieval queries, and one caption),
    spanning instruction-conditioned trajectory generation,
    language-driven semantic trajectory retrieval,
    and trajectory captioning. \looseness -1

    \item We design a \textbf{multi-dimensional evaluation protocol}
    that jointly measures trajectory fidelity, retrieval quality,
    and language groundedness, complemented by a human evaluation rubric.
    To our knowledge, this is the first protocol to assess
    language--trajectory alignment across generation, retrieval,
    and description within a single benchmark. \looseness -1

    \item We develop \textbf{Reverse Intent Reconstruction (RIR)},
    a reproducible annotation pipeline that synthesizes
    quality-controlled language annotations from real trajectories
    and is designed to be portable to new cities given compatible
    trajectory inputs and map resources. \looseness -1

    \item We develop three proof-of-concept models
    (\textbf{TrajAnchor}, \textbf{TrajFuse}, \textbf{TrajRap}),
    one per task, to calibrate the benchmark. 
    Experiments show that geometry-only baselines leave a large performance gap on language-grounded tasks, 
    confirming that TrajPrism poses a meaningful and unsaturated evaluation challenge. \looseness -1

\end{itemize}

\vspace{-10pt}

\section{Related Work}

\paragraph{Trajectory Modeling and Language Interfaces.}

Spatiotemporal trajectory modeling has matured into a rich research area~\cite{lin2024unite},
from early sequence-to-sequence similarity models~\cite{li2018deep,yao2019computing}
to recent self-supervised representation learning~\cite{chang2023contrastive,li2024t,li2025hit,zhou2025blurred,ma2024more}.
These methods operate in purely geometric or structural space,
yet comprehensive trajectory understanding requires grounding
in natural language that captures travel intent, route constraints,
and contextual semantics beyond what coordinates alone convey. \looseness -1

Large language models are increasingly applied to urban mobility tasks~\cite{han2025large},
spanning spatio-temporal prediction~\cite{li2024urbangpt}
and mobility generation and intention modeling~\cite{wang2024large,wang2026ellmob,gong2024mobility,yang2025causalmob}.
Despite this progress, trajectory modeling and language understanding
remain largely evaluated in isolation:
geometric methods are tested on trajectory benchmarks,
while language interfaces are assessed on task-specific success rates,
leaving fine-grained language--trajectory alignment on real GPS routes unaddressed.

\paragraph{Trajectory and Mobility Benchmarks.}
Traditional trajectory benchmarks such as Porto\footnote{\url{https://www.kaggle.com/competitions/pkdd-15-taxi-trip-time-prediction-ii/data}}, 
San Francisco\footnote{\url{https://ieee-dataport.org/open-access/crawdad-epflmobility}}, and 
Beijing~\cite{zheng2009mining,zheng2010geolife} focus on geometric tasks
like similarity search, travel-time estimation,
and classification.
More recently, language-aware mobility benchmarks have emerged.
TravelPlanner~\cite{xie2024travelplanner} evaluates LLM agents
on multi-day itinerary planning with tool use,
Semantic Routing~\cite{zhao2024semantic} pairs synthetic
natural-language queries with route generation,
and MobilityBench~\cite{song2026mobilitybench} tests
route-planning agents on real user queries.
However, none of these benchmarks pairs real GPS trajectories
with structured language annotations that jointly capture travel intent
and support bidirectional language--trajectory evaluation.
TrajPrism addresses this with a unified multi-task benchmark
grounded in real urban trajectories (Table~\ref{tab:related_benchmarks}).

\vspace{-10pt}

\section{TrajPrism}

\begin{figure}[t]
    \centering
    \includegraphics[width=\linewidth]{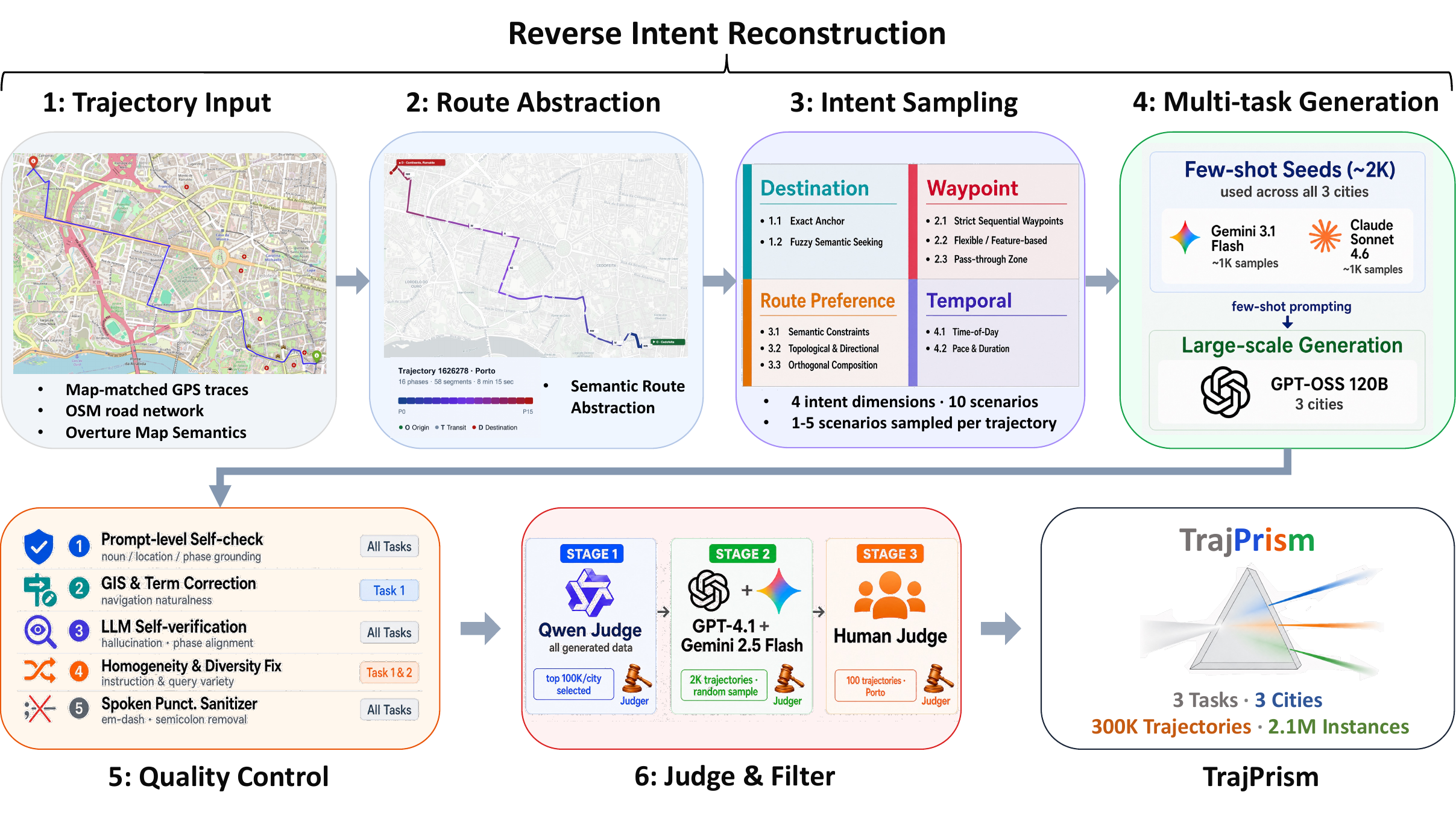}
    \caption{Dataset generation pipeline. Steps 1–4 constitute the Reverse Intent Reconstruction (RIR) framework:
    map-matched trajectories are compressed into grounded semantic phases, 
    travel intent is reconstructed via structured sampling over a four-dimensional taxonomy, 
    and multi-task language annotations are synthesized across three tasks.
    The generated data then undergoes five-stage quality control (Step 5) and cascaded LLM and human judging (Step 6), 
    yielding the final TrajPrism benchmark. \looseness -1}
    \vspace{-15pt}
    \label{fig:dataset_pipeline}
\end{figure}

TrajPrism evaluates language--trajectory alignment through three
complementary tasks: instruction-conditioned trajectory generation,
language-driven semantic trajectory retrieval, and trajectory captioning.
Each task pairs real urban trajectories with natural-language annotations,
and is assessed by a multi-dimensional evaluation protocol covering
trajectory fidelity, retrieval quality, and language groundedness.
The benchmark comprises 300K trajectories and 2.1M instances across
Porto, San Francisco, and Beijing, constructed via a reproducible
annotation pipeline (Figure~\ref{fig:dataset_pipeline}).

\begin{table}[t]
    \centering
    \small
    \caption{TrajPrism benchmark overview. Each task targets a different direction
    of language--trajectory alignment and is evaluated with complementary metrics.
    Representative metrics are listed.}
    \label{tab:task_overview}
    \setlength{\tabcolsep}{3pt}
    \renewcommand{\arraystretch}{1.15}
    \begin{tabular}{@{}p{0.22\linewidth}p{0.20\linewidth}p{0.16\linewidth}p{0.32\linewidth}@{}}
    \toprule
    \textbf{Task} & \textbf{Input} & \textbf{Output} & \textbf{Key Metrics} \\
    \midrule
    Trajectory Generation & Language + start loc. & Trajectory & Dest-Hit, Jac, DTW, Haus, EDR \\
    Trajectory Retrieval & Language query & Ranked trajs. & R@$K$, J@$K$, SR@$K$, MRR \\
    Trajectory Captioning & Trajectory & Language & BS-F1, ROUGE-L, METEOR \\
    \bottomrule
    \end{tabular}
    \vspace{-20pt}
\end{table}

\subsection{Tasks and Evaluation Protocol}

TrajPrism defines three benchmark tasks, each probing a distinct
capability for language--trajectory understanding.
Formal definitions of all evaluation metrics are provided in Appendix~\ref{sec:eval-metrics}.

\textbf{Task 1: Navigation Instruction Following.}
Given a natural language navigation instruction and a starting road segment and timestamp,
a model predicts the full trajectory (sequence of road segment IDs).
We evaluate two complementary aspects:
\textit{destination accuracy}, with Destination Hit Rate ($\text{Dest-Hit}$),
endpoint geodesic distance ($\text{Dist}$, km), and destination hit within $K$ hops ($\text{H@}K$).
For \textit{trajectory fidelity}, we measure Jaccard similarity over H3 cells ($\text{Jac}$),
Dynamic Time Warping ($\text{DTW}$, km)~\cite{keogh2005exact}, Hausdorff distance ($\text{Haus}$, km)~\cite{xie2017distributed},
and Edit Distance on Real sequences ($\text{EDR}$~\cite{chen2005robust}).

\textbf{Task 2: Trajectory Retrieval.}
Given a natural language retrieval query, a model must retrieve the
matching trajectory from a candidate pool.
We evaluate two aspects: \textit{spatial overlap}, with Jaccard at $K$ ($\text{J@}K$) 
and soft Recall ($\text{SR@}K$, the fraction of queries whose best Jaccard exceeds 0.8), 
and \textit{ranking quality}, with hard Recall ($\text{R@}K$) and Mean Reciprocal Rank ($\text{MRR}$).

\textbf{Task 3: Trajectory Captioning.}
Given a trajectory, a model must generate a factual natural language caption.
We evaluate \textit{language quality} with BERTScore F1 ($\text{BS-F1}$)~\cite{zhang2019bertscore},
ROUGE-L ($\text{R-L}$)~\cite{lin2004rouge}, and METEOR~\cite{banerjee2005meteor},
and \textit{spatial grounding} is measured by POI Recall ($\text{POI-R}$),
the proportion of ground-truth POI mentions correctly captured,
and Named Location Count ($\text{N-Loc.}$) for spatial detail coverage. \looseness -1

\subsection{Dataset Construction via Reverse Intent Reconstruction}

\subsubsection{Trajectory and Map Inputs}

TrajPrism is constructed from real map-matched urban trajectories in Porto, 
San Francisco, and Beijing. We represent each trajectory as an ordered sequence of timestamped road 
segments $\tau = ((r_1, t_1), (r_2, t_2), \ldots, (r_N, t_N))$, where 
$r_i \in \mathcal{R}$ denotes a map-matched road segment and $t_i$ is 
the corresponding timestamp.
We retrieve the road network from OpenStreetMap (OSM)%
\footnote{\url{https://www.openstreetmap.org/}} and semantic information from 
Overture Maps\footnote{\url{https://overturemaps.org/}}.
Each road segment $r_i$ is associated with a directional heading $d$ computed 
from its bearing, an H3\footnote{\url{https://h3geo.org/}} hexagonal cell $h \in \mathcal{H}$ encoding its spatial 
location, and a semantic area description $s$ derived from Overture Maps annotations 
encoding nearby POIs and urban context.
Details of retrieved information are provided in Appendix~\ref{sec:input}.

\subsubsection{Phase-based Semantic Route Abstraction}\label{sec:compression}

Following hierarchical trajectory abstraction~\cite{zhou2025blurred,li2025hit},
we compress each trajectory into semantically enriched \textit{phases}
that inline road-network structure, POI context, and area descriptions,
producing a self-contained representation an LLM can interpret
without additional lookups.
Specifically, we convert $\tau$ into a compact sequence of \textit{phases} via H3-based run-length encoding.
Each road segment is mapped to an H3 hexagonal cell through a precomputed index $\varphi: \mathcal{R} \to \mathcal{H}$. Consecutive segments sharing the same dominant H3 cell are merged into a single phase:
\begin{equation}
    P_k = \{r_i, \ldots, r_j\} \;\text{ s.t. }\; \varphi(r_l) = h_k 
    \;\forall\, r_l \in P_k
\end{equation}
This reduces a typical trajectory from hundreds of road segments to a compact sequence of mobility phases (Figure~\ref{fig:dataset_pipeline} Step 2). 
Each phase aggregates heading, duration, road names, and semantic area descriptions
 (full definition in Appendix~\ref{sec:phase-compression}).

\subsubsection{Intent Sampling via Four-Dimensional Taxonomy}

Given the compressed phase sequence from Step 2, we construct a structured intent profile for each trajectory 
through sampling over a four-dimensional intent taxonomy (Figure~\ref{fig:dataset_pipeline}, Step 3) to generate navigation instructions and retrieval queries. 
The taxonomy organizes travel intent into four dimensions (Destination, Waypoint, Route Preference, and Temporal), comprising 10 fine-grained scenarios in total.
For each trajectory, we sample $c$ scenarios without replacement, where $c \in \{1, 2, 3, 4, 5\}$ follows a distribution centered at 2–3 to reflect the natural complexity of human navigation requests. 
Single-scenario intents are retained as a minority case for short or straightforward trajectories, while composite intents combining scenarios across multiple dimensions constitute the majority. 
This sampling strategy ensures diversity in the resulting annotations while remaining grounded in the spatial and semantic characteristics of each trajectory. 
More details are shown in Appendix~\ref{sec:intent-sampling}. 

The sampled scenario set is then passed to the generation stage as a structured prompt constraint, 
ensuring all generated annotations faithfully reflect the inferred intent rather than generic route descriptions. \looseness -1

\subsubsection{Multi-task Generation}

Given the compressed phase sequence $\{\mathbf{p}_k\}_{k=1}^K$ and sampled intent profile from 
Steps 2--3, we generate language annotations across three tasks via large 
language models (Figure~\ref{fig:dataset_pipeline}, Step 4). 
Each task defines a distinct input-output format grounded in the same underlying trajectory.

\textbf{Task 1: Navigation Instruction} generates three stylistic variants (Literal, Concise, and Chatty) per trajectory, 
conditioned on the sampled intent scenarios. 
To maximize lexical and tonal diversity, each variant is additionally conditioned on a speaker persona, sentence-form hint, 
and length guidance (Appendix~\ref{sec:data_scenarios}).

\textbf{Task 2: Trajectory Retrieval} generates three retrieval queries per trajectory, 
collectively spanning all four intent dimensions to ensure broad semantic coverage.

\textbf{Task 3: Trajectory Captioning} generates a single third-person, 
factual caption per trajectory following a fixed narrative structure: origin, 
key route behaviors and semantic zones traversed, and destination. The model 
is instructed to adopt an analyst's perspective, describing what the trajectory 
\textit{did} on the map without inferring driver intent.

Each trajectory thus yields seven instances ($3 + 3 + 1$).
To balance quality and scale, we adopt a two-stage generation strategy: 
approximately 2K few-shot seed examples are first produced using Gemini 3.1 Flash-Lite~\footnote{\url{https://ai.google.dev/gemini-api/docs/models/gemini-3.1-flash-lite-preview}}
and Claude Sonnet 4.6~\footnote{\url{https://www.anthropic.com/claude/sonnet}}, then used as in-context demonstrations for large-scale 
generation with GPT-OSS 120B~\cite{agarwal2025gpt} across all three cities. 

\subsection{Quality Control and Data Judging}

To ensure the resulting annotations meet benchmark-level reliability, 
we apply a two-phase verification pipeline 
(Figure~\ref{fig:dataset_pipeline}, Steps 5--6). 
The first phase applies five deterministic quality control stages
covering noun/location/phase grounding, GIS terminology correction,
LLM-based hallucination verification, lexical diversity enforcement,
and punctuation sanitization. \looseness -1

Drawing on the LLM-as-a-judge paradigm~\cite{zheng2023judging}, 
we adopt a cascaded judging strategy that combines scalable LLM scoring, cross-model validation, 
and human verification in the second phase, to ensure annotation quality is both measurable and reproducible across all three cities.
A Qwen-based judge~\cite{qwen3.5} 
first scores all generated data and selects the top 100K trajectories per city, 
a sample of 2K trajectories is then independently re-evaluated by GPT-4.1\footnote{\url{https://developers.openai.com/api/docs/models/gpt-4.1}} and 
Gemini 2.5 Flash~\cite{comanici2025gemini} to validate scoring consistency, and 100 randomly sampled Porto trajectories 
are assessed by human annotators to establish a quality baseline.
All LLM and human judges share the same evaluation rubric, enabling direct comparison and inter-annotator agreement analysis.
The evaluation rubric for both LLM and human judges is detailed in Appendix~\ref{sec:dataset-judging}.
Human annotators achieve 91.7\% mean $\pm$1 agreement among themselves; human--LLM cross-agreement reaches 92.9\%, 
confirming that LLM judges reliably approximate human evaluation (Tables~\ref{tab:human_mean_score} and~\ref{tab:human_llm_agreement}).

\section{Experiments}\label{sec:experiments}

\begin{figure}[t]
    \centering
    \includegraphics[width=\linewidth]{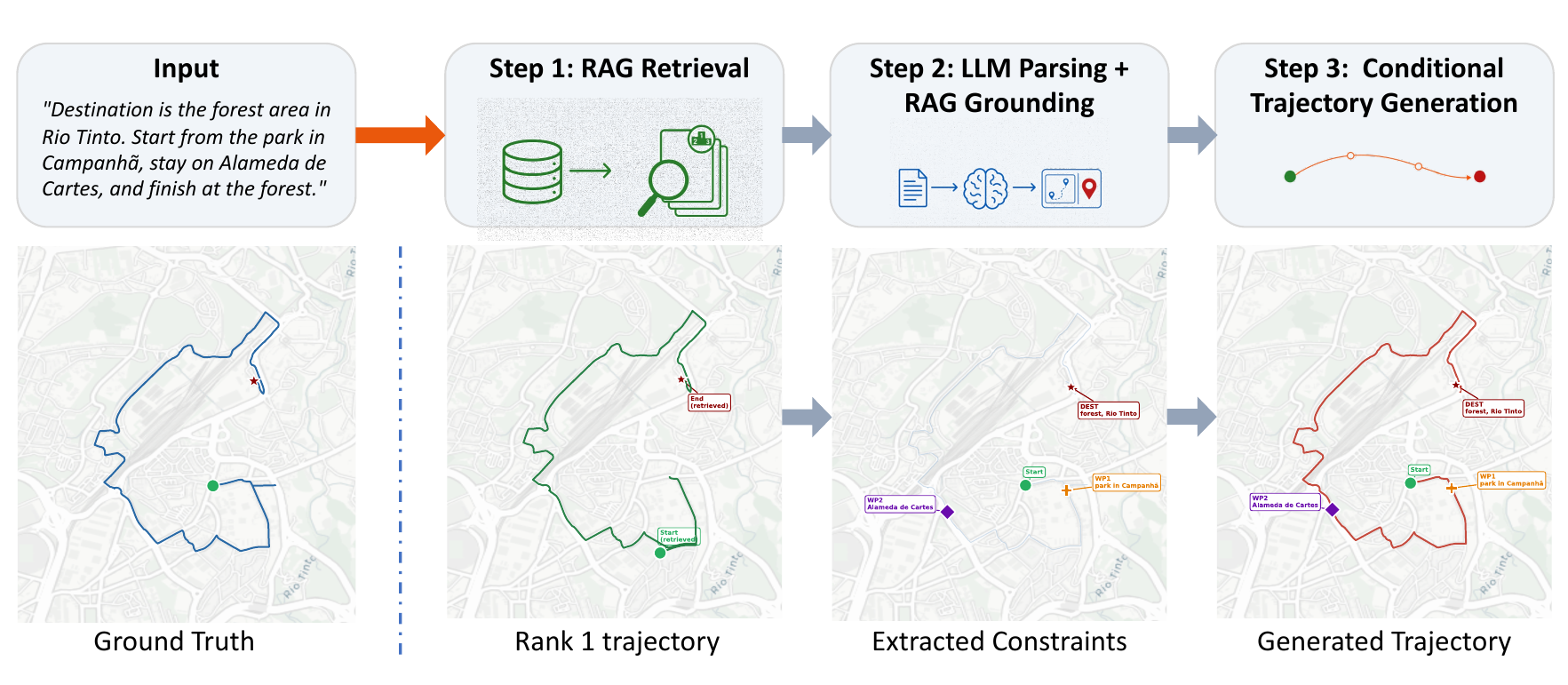}
    \caption{\textbf{TrajAnchor pipeline (Task~1).}
    Step~1 retrieves similar training trajectories;
    Step~2 extracts and grounds spatial constraints via LLM;
    Step~3 generates the route via chain Dijkstra.
    Bottom: Porto example with ground truth on the left.}
    \vspace{-15pt}
    \label{fig:trajanchor-pipeline}
\end{figure}

\begin{table*}[t]
    \centering
    \small
    \caption{Task 1: Instruction-conditioned Trajectory Generation results.
    Best per dataset in \textbf{bold}, second best \underline{underlined}.
    $^\dagger$: Qwen3.5-4B (default); $^\star$: Claude Sonnet~4.6.}
    \label{tab:task1_baselines}
    \setlength{\tabcolsep}{4.5pt}
    \begin{tabular}{llccccccc}
    \toprule\toprule
    \textbf{Dataset} & \textbf{Method} 
      & \textbf{H@5}$\uparrow$ 
      & \textbf{Dest-Hit}$\uparrow$ 
      & \textbf{Dist(km)}$\downarrow$ 
      & \textbf{Jac}$\uparrow$ 
      & \textbf{DTW}$\downarrow$ 
      & \textbf{Haus}$\downarrow$ 
      & \textbf{EDR}$\downarrow$ \\
    \midrule\midrule
    \multirow{7}{*}{Porto}
      & DestSP  (BM25)     & 0.0823 & 0.1357 & 2.4099 & 0.2454 & 55.2871 & 2.0800 & 0.7814 \\
      & DestSP (Embed)     & 0.0955 & 0.1533 & 2.0435 & 0.2707 & 47.0833 & 1.8446 & 0.7605 \\
      \addlinespace[3pt]
      & ConstrSP           & 0.0912 & 0.1480 & 2.1937 & 0.2188 & 63.0137 & 2.0989 & 0.8280 \\
      \addlinespace[3pt]
      & TrajAnchor (Q3.5-2B) & 0.1003 & 0.1578 & 1.8201 & 0.2855 & 42.0668 & 1.7106 & 0.7532 \\
      & \textbf{TrajAnchor}$^\dagger$ & 0.0987 & 0.1540 & \underline{1.6723} & 0.2955 & \underline{38.6676} & 1.6193 & 0.7463 \\
      & \textbf{TrajAnchor}$^\star$ (CS4.6) & \textbf{0.1104} & \textbf{0.1701} & 1.6728 & \textbf{0.3013} & 38.6969 & \underline{1.6060} & \textbf{0.7386} \\
      & TrajAnchor (Q3.5-9B) & \underline{0.1018} & \underline{0.1595} & \textbf{1.6333} & \underline{0.3003} & \textbf{37.5961} & \textbf{1.5912} & \underline{0.7420} \\
      \midrule
    \multirow{7}{*}{SF}
      & DestSP (BM25)      & 0.1369 & 0.1476 & 2.6262 & 0.2084 & 53.8048 & 2.1931 & 0.8317 \\
      & DestSP (Embed)     & 0.1644 & 0.1771 & 2.4647 & 0.2245 & 51.7005 & 2.1082 & 0.8198 \\
      \addlinespace[3pt]
      & ConstrSP           & 0.1590 & 0.1747 & 2.3691 & 0.1942 & 58.2234 & 2.2361 & 0.8652 \\
      \addlinespace[3pt]
      & TrajAnchor (Q3.5-2B) & \underline{0.1756} & 0.1948 & 2.3053 & 0.2341 & 48.5685 & 2.0084 & 0.8135\\
      & \textbf{TrajAnchor}$^\dagger$ & 0.1743 & \underline{0.1956} & \underline{2.2167} & \underline{0.2405} & \underline{46.8433} & \underline{1.9574} & \underline{0.8090} \\
      & \textbf{TrajAnchor}$^\star$ (CS4.6) & \textbf{0.1837} & \textbf{0.2041} & \textbf{2.2070} & \textbf{0.2457} & \textbf{46.0980} & \textbf{1.9403} & \textbf{0.8037} \\
      & TrajAnchor (Q3.5-9B) & 0.1754 & \underline{0.1956} & 2.2633 & 0.2371 & 47.6012 & 1.9716 & 0.8112 \\
    \midrule
    \multirow{7}{*}{BJ}
      & DestSP (BM25)      & 0.0916 & 0.0967 & 4.3284 & 0.2075 & 76.1076 & 3.6399 & 0.7938 \\
      & DestSP (Embed)     & 0.1501 & 0.1611 & 3.8366 & 0.2489 & 67.7288 & 3.2812 & 0.7557 \\
      \addlinespace[3pt]
      & ConstrSP           & 0.1530 & 0.1570 & \textbf{3.5433} & 0.2127 & 72.2267 & 3.3429 & 0.8090 \\
      \addlinespace[3pt]
      & TrajAnchor (Q3.5-2B) & 0.1549 & 0.1666 & 3.7530 & 0.2559 & 66.0261 & 3.2279 & 0.7495 \\
      & \textbf{TrajAnchor}$^\dagger$ & 0.1634 & 0.1742 & 3.7121 & 0.2605 & 65.1863 & 3.1908 & 0.7447 \\
      & \textbf{TrajAnchor}$^\star$ (CS4.6) & \textbf{0.1800} & \textbf{0.1945} & \underline{3.5931} & \textbf{0.2709} & \textbf{63.1457} & \textbf{3.0992} & \textbf{0.7345} \\
      & TrajAnchor (Q3.5-9B) & \underline{0.1666} & \underline{0.1808} & 3.6891 & \underline{0.2635} & \underline{64.6283} & \underline{3.1714} & \underline{0.7415} \\
    \bottomrule\bottomrule
  \end{tabular}
  \vspace{-20pt}
\end{table*}

\begin{table*}[ht]
  \centering
  \small
  \setlength{\tabcolsep}{4pt}
  \caption{Task 2: Language-driven Semantic Trajectory Retrieval results. Best per dataset in \textbf{bold}, second best \underline{underlined}.\looseness -1}
  \label{tab:task2_retrieval}
  \begin{tabular}{llccccccc}
  \toprule
  \toprule
  \textbf{Dataset} & \textbf{Method} & \textbf{J@1} & \textbf{J@5} & \textbf{SR@1} & \textbf{SR@5} & \textbf{R@1} & \textbf{R@10} & \textbf{MRR} \\
  \midrule
  \midrule
  \multirow{10}{*}{Porto}
    & TF-IDF + SVD          & 0.2381 & 0.4648 & 0.0821 & 0.2009 & 0.0245 & 0.1218 & 0.0490 \\
    & BoW + SVD             & 0.2368 & 0.4632 & 0.0811 & 0.2003 & 0.0252 & 0.1220 & 0.0493 \\
  \cmidrule(lr){2-9}
    & UniTraj          & 0.1655 & 0.3383 & 0.0598 & 0.1898 & 0.0154 & 0.0814 & 0.0319 \\
    & TrajCL           & \underline{0.2783} & 0.5239 & \underline{0.1526} & 0.3266 & 0.0839 & 0.2798 & 0.1376 \\
    & T-JEPA           & 0.2544 & 0.4806 & 0.1236 & 0.2712 & 0.0619 & 0.2245 & 0.1050 \\
    & BLUE             & 0.2060 & 0.4139 & 0.0780 & 0.1867 & 0.0295 & 0.1391 & 0.0573 \\
  \cmidrule(lr){2-9}
    &    Sem. (Zero-shot)   & 0.0955 & 0.2087 & 0.0420 & 0.0950 & 0.0315 & 0.1105 & 0.0523 \\
    &    Sem.               & 0.2621 & \underline{0.5273} & 0.1489 & \underline{0.3462} & \underline{0.0940} & \textbf{0.3185} & \textbf{0.1548} \\
    & Road GATv2            & 0.2343 & 0.4561 & 0.0870 & 0.2059 & 0.0357 & 0.1547 & 0.0660 \\
    & \textbf{TrajFuse}  & \textbf{0.2943} & \textbf{0.5443} & \textbf{0.1631} & \textbf{0.3471} & \textbf{0.0957} & \underline{0.3088} & \underline{0.1538} \\
  \midrule\midrule
  \multirow{10}{*}{SF}
    & TF-IDF + SVD          & 0.1666 & 0.3436 & 0.0415 & 0.1070 & 0.0276 & 0.1182 & 0.0507 \\
    & BoW + SVD             & 0.1634 & 0.3423 & 0.0393 & 0.1051 & 0.0274 & 0.1130 & 0.0493 \\
  \cmidrule(lr){2-9}
    & UniTraj          & 0.0542 & 0.1449 & 0.0117 & 0.0358 & 0.0067 & 0.0362 & 0.0142 \\
    & TrajCL           & \underline{0.2863} & \underline{0.5090} & \underline{0.2163} & \underline{0.3926} & \underline{0.1983} & 0.4424 & \underline{0.2718} \\
    & T-JEPA           & 0.1628 & 0.3488 & 0.0941 & 0.2113 & 0.0804 & 0.2422 & 0.1265 \\
    & BLUE             & 0.1476 & 0.3291 & 0.0777 & 0.1870 & 0.0653 & 0.2192 & 0.1074 \\
  \cmidrule(lr){2-9}
    &    Sem. (Zero-shot)   & 0.1153 & 0.2238 & 0.0850 & 0.1557 & 0.0819 & 0.1843 & 0.1121 \\
    &    Sem.               & 0.2421 & 0.4882 & 0.1858 & 0.3841 & 0.1726 & \underline{0.4508} & 0.2543 \\
    & Road GATv2            & 0.1590 & 0.3424 & 0.0592 & 0.1475 & 0.0475 & 0.1738 & 0.0804 \\
    & \textbf{TrajFuse}  & \textbf{0.3085} & \textbf{0.5454} & \textbf{0.2397} & \textbf{0.4345} & \textbf{0.2233} & \textbf{0.4837} & \textbf{0.3030} \\
  \midrule\midrule
  \multirow{10}{*}{BJ}
    & TF-IDF + SVD          & 0.0970 & 0.2508 & 0.0203 & 0.0700 & 0.0163 & 0.0938 & 0.0350 \\
    & BoW + SVD             & 0.0928 & 0.2428 & 0.0180 & 0.0644 & 0.0147 & 0.0888 & 0.0323 \\
  \cmidrule(lr){2-9}
    & UniTraj          & 0.0190 & 0.0500 & 0.0068 & 0.0171 & 0.0041 & 0.0199 & 0.0080 \\
    & TrajCL           & \underline{0.2158} & \underline{0.4325} & 0.1652 & 0.3363 & 0.1562 & 0.4076 & 0.2278 \\
    & T-JEPA           & 0.1602 & 0.3497 & 0.1137 & 0.2515 & 0.1055 & 0.3185 & 0.1637 \\
    & BLUE             & 0.0790 & 0.2187 & 0.0301 & 0.0953 & 0.0257 & 0.1321 & 0.0523 \\
  \cmidrule(lr){2-9}
    &    Sem. (Zero-shot)   & 0.1443 & 0.2500 & 0.1277 & 0.2128 & 0.1255 & 0.2547 & 0.1616 \\
    &    Sem.               & 0.2130 & 0.4272 & \underline{0.1802} & \underline{0.3556} & \underline{0.1744} & \underline{0.4344} & \underline{0.2480} \\
    & Road GATv2            & 0.1261 & 0.3039 & 0.0604 & 0.1598 & 0.0549 & 0.2099 & 0.0957 \\
    & \textbf{TrajFuse}  & \textbf{0.2653} & \textbf{0.4961} & \textbf{0.2213} & \textbf{0.4108} & \textbf{0.2138} & \textbf{0.4896} & \textbf{0.2929} \\
  \bottomrule
  \bottomrule
  \end{tabular}
  \vspace{-20pt}
\end{table*}
  
\begin{table}[t]
  \centering
  \small
  \setlength{\tabcolsep}{4pt}
  \caption{Task 3: Trajectory Captioning results. Best per dataset in \textbf{bold}, second best \underline{underlined}.}
  \label{tab:task3_results}
  \begin{tabular}{llcccccccc}
  \toprule
  \textbf{Dataset} & \textbf{Method} & \textbf{BS-F1} & \textbf{ROUGE-L} & \textbf{METEOR} & \textbf{POI-R} & \textbf{N-Loc.} & \textbf{In} & \textbf{Out} & \textbf{Lat.(s)} \\
  \midrule
  \multirow{4}{*}{Porto}
    & Struct.       & 0.854 & 0.245 & 0.206 & 0.121 & 4.69 & 477  & 68  & 1.18 \\
    & Sem.          & 0.864 & 0.259 & 0.227 & \textbf{0.408} & 5.46 & 2686 & 72  & 2.16 \\
    & Sem. Distill. & \underline{0.871} & \underline{0.310} & \textbf{0.315} & 0.033 & 5.30 & 468  & 119 & 1.81 \\
    & \textbf{TrajRap}      & \textbf{0.885} & \textbf{0.321} & \underline{0.314} & \underline{0.341} & 4.85 & 5025 & 93 & 5.89 \\
  \midrule
  \multirow{4}{*}{SF}
    & Struct.       & 0.867 & 0.227 & 0.173 & 0.015 & 5.32 & 411  & 45  & 0.87 \\
    & Sem.          & 0.868 & 0.237 & 0.200 & \textbf{0.095} & 6.48 & 2419 & 54  & 1.83 \\
    & Sem. Distill. & \underline{0.881} & \textbf{0.316} & \textbf{0.313} & 0.017 & 5.73 & 401  & 101 & 1.56 \\
    & \textbf{TrajRap}       & \textbf{0.885} & \underline{0.307} & \underline{0.298} & \underline{0.087} & 5.99 & 4703 & 82 & 3.14 \\
  \midrule
  \multirow{4}{*}{BJ}
    & Struct.       & 0.836 & 0.167 & 0.127 & 0.238 & 2.81 & 486  & 50  & 0.96 \\
    & Sem.          & 0.849 & 0.210 & 0.183 & \underline{0.284} & 5.54 & 1594 & 68  & 1.60 \\
    & Sem. Distill. & \underline{0.856} & \underline{0.259} & \underline{0.233} & 0.234 & 2.44 & 477  & 91  & 1.47 \\
    & \textbf{TrajRap}       & \textbf{0.876} & \textbf{0.308} & \textbf{0.293} & \textbf{0.287} & 4.74 & 4399 & 90 & 3.12 \\
  \bottomrule
  \end{tabular}
  \vspace{-20pt}
\end{table}

\subsection{Proof-of-Concept Models}

We develop one proof-of-concept model per task to instantiate
the benchmark and calibrate its difficulty, prioritizing coverage
of the three task interfaces over architectural novelty.
\textbf{TrajAnchor} (Task~1) uses a retrieved training trajectory
as a spatial \textit{anchor} for route generation.
\textbf{TrajFuse} (Task~2) \textit{fuses} geometric and semantic
trajectory representations for cross-modal contrastive retrieval.
\textbf{TrajRap} (\textbf{R}etrieval-\textbf{A}ugmented
\textbf{P}rofiling, for Task~3) generates factual trajectory captions
from retrieved semantic context.
We detail TrajAnchor below as the most involved pipeline;
TrajFuse and TrajRap are described in Appendix~\ref{sec:poc-details}.

\textbf{TrajAnchor} (Task~1) is a three-stage pipeline for
instruction-conditioned trajectory generation
(Figure~\ref{fig:trajanchor-pipeline}).
\textit{Step~1 (Retrieval):} given a navigation instruction, we encode it
with a text embedding model~\cite{nussbaum2024nomic} and retrieve the most similar training
trajectory via cosine similarity over a pre-built index, 
optionally filtered to trajectories sharing the same starting H3 cell,
followed by destination-proximity reranking.
\textit{Step~2 (Constraint Extraction):} an LLM parses the instruction
to extract structured spatial constraints (destination, waypoints,
route preferences). The destination phrase is grounded to specific road segments
by matching against H3 cell area descriptions via embedding similarity.
\textit{Step~3 (Route Generation):} the extracted constraints are
combined with skeleton waypoints sampled from the retrieved trajectory
to seed a soft-weighted chain Dijkstra search over the road network,
producing the final predicted trajectory.
We experiment with multiple LLM backbones
(Qwen-3.5 2B/4B/9B, Claude Sonnet~4.6) for constraint extraction,
and all variants share the same retrieval and routing modules.
Unless otherwise specified, 
all baselines and our TrajAnchor use the Qwen-3.5 4B backbone by default. \looseness -1

\subsection{Results}

\paragraph{Setup.}
We split each city into train, val, and test splits, 
each containing 70\%, 10\%, and 20\% of the trajectories, respectively.
This yields 210k, 30k, and 60k navigation instructions and retrieval queries for each city, respectively;
and 70k, 10k, and 20k trajectory caption for each city, respectively.
We evaluate on the test split of each city (Porto, San Francisco, Beijing).
For Task~1, we compare TrajAnchor against two geometry-only baselines:
\textit{DestSP}, which only routes to the LLM-extracted destination via shortest path, using either BM25 or embedding-based destination retrieval,
and \textit{ConstrSP}, which adds additional LLM-extracted waypoint constraints.
For Task~2, we compare TrajFuse against
text-based baselines (TF-IDF + SVD, BoW + SVD),
four trajectory encoders
(UniTraj~\cite{zhu2024unitraj}, TrajCL~\cite{chang2023contrastive},
T-JEPA~\cite{li2024t}, BLUE~\cite{zhou2025blurred}) under fine-tuned settings,
an H3 semantic encoder that represents trajectories by their cell-level area descriptions
(with and without fine-tuning),
and a Road GATv2 encoder that learns from trajectory graph structure.
For Task~3, we compare TrajRap against three ablations:
a structural-only baseline where an LLM receives only trajectory coordinates (Struct.),
a semantic-augmented variant (Sem.) where an LLM receives additional semantic context,
and a distilled variant (Sem.\ Distill.) that is LoRA~\cite{hu2022lora} fine-tuned with semantic context
but receives only structural input at inference, testing whether the model
internalizes spatial semantics through training.
Implementation details (hyperparameters, training schedules)
are provided in Appendix~\ref{sec:poc-details}.
All models share a common spatial index of H3 hexagonal cells at resolution~9,
each annotated with a natural-language area description covering POIs,
land use, and road context, encoded with Nomic Embed~\cite{nussbaum2024nomic}
(details in Appendix~\ref{sec:input}).
All experiments are conducted on a node with 4$\times$ NVIDIA H100 80GB GPUs.
Tables~\ref{tab:task1_baselines}--\ref{tab:task3_results} report results
across the three tasks and three cities.
We highlight benchmark-level insights below, and
per-method details and additional ablations are in
Appendix~\ref{sec:poc-details}.

\textbf{Task~1: Navigation Instruction Following}
(Table~\ref{tab:task1_baselines}).
Even the best TrajAnchor variant achieves only 17--20\% Dest-Hit
and $\sim$0.25--0.30 Jaccard across cities,
indicating that the majority of generated trajectories still
deviate from the ground truth.
Language-augmented TrajAnchor consistently outperforms the
geometry-only baselines (DestSP, ConstrSP) on trajectory-level
metrics (Jac, DTW, Haus),
confirming that leveraging instruction semantics beyond
destination extraction improves route fidelity.
ConstrSP reaches approximately correct destinations (competitive Dist)
but fails to reproduce the intended route shape (low Jac, high DTW).
Beijing proves the most challenging setting,
with the highest DTW and Hausdorff values across all methods.
Scaling the LLM backbone from Qwen3.5-4B to Claude Sonnet~4.6
yields consistent gains across all metrics,
suggesting that stronger language understanding
directly benefits constraint extraction and route fidelity. 
Additional destination-proximity metrics (H@$K$) and routing
diagnostics are reported in Table~\ref{tab:task1_supp};
an oracle analysis revealing the retrieval ceiling is given
in Table~\ref{tab:task1_oracle} (Appendix).
and qualitative examples illustrating good, moderate, 
and poor predictions are shown in Figure~\ref{fig:task1-trio} (Appendix).\looseness -1

\textbf{Task~2: Trajectory Retrieval}
(Table~\ref{tab:task2_retrieval}).
Fine-tuned trajectory encoders that lack semantic input
show highly uneven performance:
TrajCL is competitive, but UniTraj
nearly fails (R@1 $<$ 0.02 on all cities),
demonstrating that geometric trajectory representations
alone do not reliably capture language-expressed intent.
Methods incorporating H3 semantic descriptions
consistently outperform pure trajectory encoders,
and TrajFuse, which fuses geometric and semantic
representations, achieves the best spatial overlap
(J@1, SR@5) across all three cities.
Nevertheless, even TrajFuse's R@1 remains below 0.10
in Porto and below 0.23 elsewhere,
leaving substantial room for improvement
in exact trajectory identification from natural language. 
Additional retrieval metrics (J@10, SR@10, R@50) are reported in
Table~\ref{tab:task2_supp}, and a per-intent-dimension breakdown
is shown in Figure~\ref{fig:retrieval-radar} (Appendix).\looseness -1

\textbf{Task~3: Trajectory Captioning}
(Table~\ref{tab:task3_results}).
BS-F1 remains consistently high (>0.83) across all methods,
indicating that fluency is not the bottleneck;
the key challenge lies in grounding captions with accurate spatial references (POI-R).
The structural-only baseline produces fluent text but
achieves very low POI Recall (as low as 0.015 in SF),
indicating that without access to semantic context
the LLM cannot ground its descriptions in real place names.
Adding H3 semantic descriptions (Sem.) substantially
improves grounding (POI-R jumps from 0.12 to 0.41 in Porto),
while knowledge distillation (Sem.\ Distill.) and
the full TrajRap pipeline further improve language metrics
(ROUGE-L, METEOR) while maintaining reasonable spatial grounding.
Notably, Sem.\ Distill.\ achieves strong ROUGE-L and METEOR
despite receiving no semantic input at inference,
suggesting that spatial semantics can be partially internalized
through fine-tuning, though POI Recall drops sharply
without direct access to area descriptions.
Across all settings, ROUGE-L stays below 0.33 and POI-R
remains far from 1.0,
indicating that faithful, well-grounded trajectory captioning
is far from solved. \looseness -1

\vspace{-6pt}

\paragraph{Overall takeaway.}
We distill four cross-task findings from the results above
and the supplementary analyses in the Appendix.

\textit{(1) Language consistently helps, but the task is far from solved.}
Across all three tasks, methods that leverage natural-language
semantics outperform geometry-only or structure-only baselines.
Yet absolute performance remains low on every metric
(Dest-Hit $\leq$ 0.20, R@1 $\leq$ 0.23, POI-R $\ll$ 1.0),
confirming that TrajPrism poses a meaningful and unsaturated
evaluation challenge.

\textit{(2) Each task has a distinct bottleneck.}
Generation is limited primarily by destination selection:
oracle analysis (Table~\ref{tab:task1_oracle}) shows that
Jaccard nearly doubles when the correct destination is given,
while the routing algorithm itself is not the main source of error.
Retrieval struggles most with temporal and preference-based intents,
whereas spatially unambiguous queries are better handled
(Figure~\ref{fig:retrieval-radar}).
Captioning saturates on fluency (BS-F1 $>$ 0.83) but fails on
spatial grounding, with POI Recall far below 1.0 across all methods.

\textit{(3) Map-level semantic context is a shared enabler.}
Incorporating H3 area descriptions yields consistent gains on all
three tasks: improving constraint grounding in generation,
enabling cross-modal fusion in retrieval (TrajFuse leads on J@1
and SR@5), and boosting POI Recall from 0.12 to 0.41 in captioning.
This suggests that structured spatial knowledge is a key ingredient
for bridging geometry and language.

\textit{(4) Difficulty varies across cities and instruction styles.}
Beijing is hardest for generation due to longer trajectories and
larger road networks.
Porto poses the greatest retrieval challenge (lowest R@1).
Literal instructions consistently yield higher fidelity than
Concise and Chatty variants (Figure~\ref{fig:task1-style-bars}),
These patterns highlight that language-grounded trajectory
understanding requires models to jointly handle spatial,
temporal, and linguistic complexity across diverse urban settings. \looseness -1

\vspace{-6pt}

\section{Conclusion}

We presented TrajPrism, a multi-task benchmark for evaluating language--trajectory alignment on real urban GPS trajectories.
By unifying generation, retrieval, and captioning in one framework,
TrajPrism enables joint assessment of trajectory fidelity, retrieval quality, and language groundedness on the same observed trips.
Proof-of-concept experiments confirm that current models leave substantial headroom on all tasks, establishing language--trajectory alignment as an open challenge.
We release the benchmark, code, and a reproducible annotation pipeline portable across cities. 
We hope TrajPrism serves as a foundation for advancing language-grounded urban trajectory understanding. \looseness -1

\begin{ack}
  We express our gratitude to Sharon AI for providing access to NVIDIA H100 GPUs.
  We acknowledge the resources and services from the National Computational Infrastructure (NCI), which is supported by the Australian Government.
\end{ack}

\newpage

\bibliographystyle{abbrv}
\bibliography{ref}

\newpage


\appendix

\section{Appendix}\label{sec:appendix}

\subsection{Evaluation Metrics}\label{sec:eval-metrics}

We provide formal definitions of all evaluation metrics used in TrajPrism.

\subsubsection{Task 1: Navigation Instruction Following}

\paragraph{Destination Accuracy.}
\begin{itemize}[leftmargin=*,itemsep=2pt]
    \item \textbf{Destination Hit Rate (Dest-Hit)}: fraction of predictions
    whose final H3 cell matches the ground-truth destination.
    \item \textbf{Endpoint Distance (Dist)}: geodesic distance (km) between
    the predicted and ground-truth endpoints.
    \item \textbf{Destination Hit within $K$ Hops (H@$K$)}: fraction of
    predictions whose endpoint is within $K$ road-segment hops of the
    ground-truth destination.
\end{itemize}

\paragraph{Trajectory Fidelity.}
\begin{itemize}[leftmargin=*,itemsep=2pt]
    \item \textbf{Jaccard (Jac)}: Jaccard similarity over the sets of H3 cells
    traversed by the predicted and ground-truth trajectories.
    \item \textbf{Dynamic Time Warping (DTW)}~\cite{keogh2005exact}:
    DTW distance (km) between the GPS coordinate sequences.
    \item \textbf{Hausdorff (Haus)}~\cite{xie2017distributed}:
    Hausdorff distance (km) measuring the worst-case spatial deviation.
    \item \textbf{Edit Distance on Real sequences (EDR)}~\cite{chen2005robust}:
    normalized edit distance under a spatial threshold.
\end{itemize}

\paragraph{Oracle Metrics (O-prefix).}
Retrieval-based routing methods first retrieve multiple destination
candidates together with associated spatial constraints (waypoints,
route preferences), then generate a route for each candidate and
select the final prediction.
To disentangle destination-selection errors from route-planning
errors, we introduce \emph{oracle} variants of the above metrics:
for each query we generate routes for all retrieved candidates and
select the one whose endpoint is closest to the ground-truth
destination, then evaluate that route.
The oracle metrics (O-H@5, O-Dest, O-Dist, O-Jac, O-DTW, O-Haus,
O-EDR) share the same definitions as their standard counterparts
but reveal the performance ceiling achievable if the best candidate
were always chosen, thereby isolating the routing stage from
the candidate selection stage.

\subsubsection{Task 2: Trajectory Retrieval}

\paragraph{Spatial Overlap.}
\begin{itemize}[leftmargin=*,itemsep=2pt]
    \item \textbf{Jaccard at $K$ (J@$K$)}: mean Jaccard similarity between
    the top-$K$ retrieved trajectories and the ground truth.
    \item \textbf{Soft Recall at $K$ (SR@$K$)}: fraction of queries for which
    at least one of the top-$K$ retrievals has Jaccard $> 0.8$ with the
    ground truth.
\end{itemize}

\paragraph{Ranking Quality.}
\begin{itemize}[leftmargin=*,itemsep=2pt]
    \item \textbf{Hard Recall at $K$ (R@$K$)}: fraction of queries for which
    the ground-truth trajectory appears in the top-$K$ results.
    \item \textbf{Mean Reciprocal Rank (MRR)}: mean of the reciprocal rank of
    the first correct result across all queries.
\end{itemize}

\subsubsection{Task 3: Trajectory Captioning}

\paragraph{Language Quality.}
\begin{itemize}[leftmargin=*,itemsep=2pt]
    \item \textbf{BERTScore F1 (BS-F1)}~\cite{zhang2019bertscore}:
    token-level semantic similarity between predicted and reference captions.
    \item \textbf{ROUGE-L (R-L)}~\cite{lin2004rouge}:
    longest common subsequence overlap.
    \item \textbf{METEOR}~\cite{banerjee2005meteor}:
    alignment-based metric accounting for synonyms and stemming.
\end{itemize}

\paragraph{Spatial Grounding.}
\begin{itemize}[leftmargin=*,itemsep=2pt]
    \item \textbf{POI Recall (POI-R)}: proportion of ground-truth POI mentions
    correctly captured in the generated caption.
    \item \textbf{Named Location Count (N-Loc.)}: number of distinct named
    locations mentioned, measuring spatial detail coverage.
\end{itemize}

\subsection{Proof-of-Concept Models: TrajFuse and TrajRap}\label{sec:poc-details}

\begin{figure}[htbp]
    \centering
    \includegraphics[width=\linewidth]{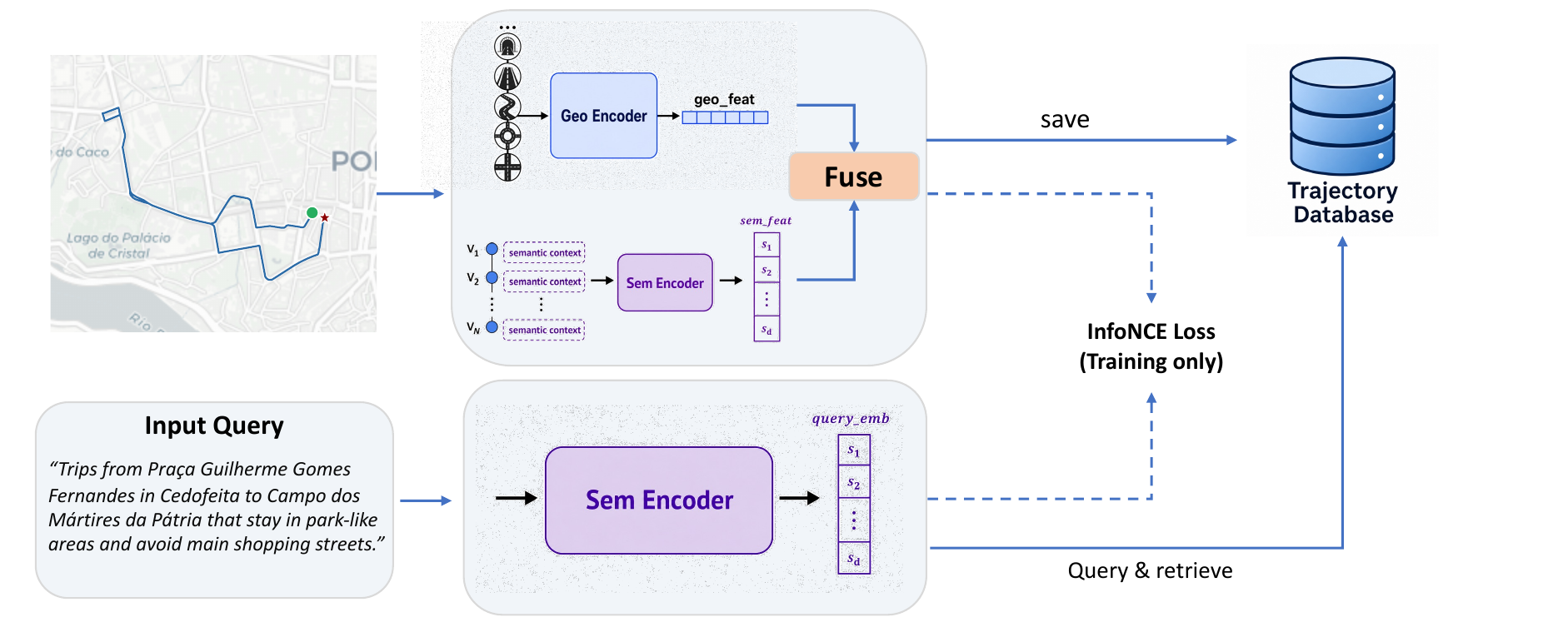}
    \caption{\textbf{TrajFuse architecture (Task~2).}
    A dual-encoder framework fuses geometric trajectory embeddings
    (from a fine-tuned TrajCL encoder) with H3-cell semantic embeddings
    and aligns them with text query embeddings via contrastive learning
    for cross-modal trajectory retrieval.}
    \label{fig:trajfuse-architecture}
\end{figure}

\textbf{TrajFuse} (Task~2) addresses language-driven trajectory
retrieval via a dual-encoder contrastive framework
(Figure~\ref{fig:trajfuse-architecture}).
On the \emph{trajectory side}, each route is represented through
two complementary branches:
(i)~a \textbf{Geo Encoder} (TrajCL~\cite{chang2023contrastive})
that takes the raw trajectory point sequence and produces a
geometric embedding \texttt{geo\_feat} capturing spatial shape and
sequencing;
(ii)~a \textbf{Sem Encoder} that encodes the H3-cell semantic
context (POIs, land use, district descriptions) along each
visited cell $V_1, \dots, V_N$ into a semantic embedding
\texttt{sem\_feat} using the same text
encoder~\cite{nussbaum2024nomic} as the query side.
The two embeddings are combined by a \textbf{Fuse} module
(concat + linear projection) into a single trajectory
representation, which is saved to a trajectory database.
On the \emph{query side}, the retrieval query is encoded by
the shared text encoder into \texttt{query\_emb}.
During training, an InfoNCE loss with a learnable temperature
aligns the fused trajectory embeddings with the corresponding
query embeddings; at inference, retrieval is performed via cosine
similarity between \texttt{query\_emb} and the trajectory database.

\begin{figure}[htbp]
    \centering
    \includegraphics[width=\linewidth]{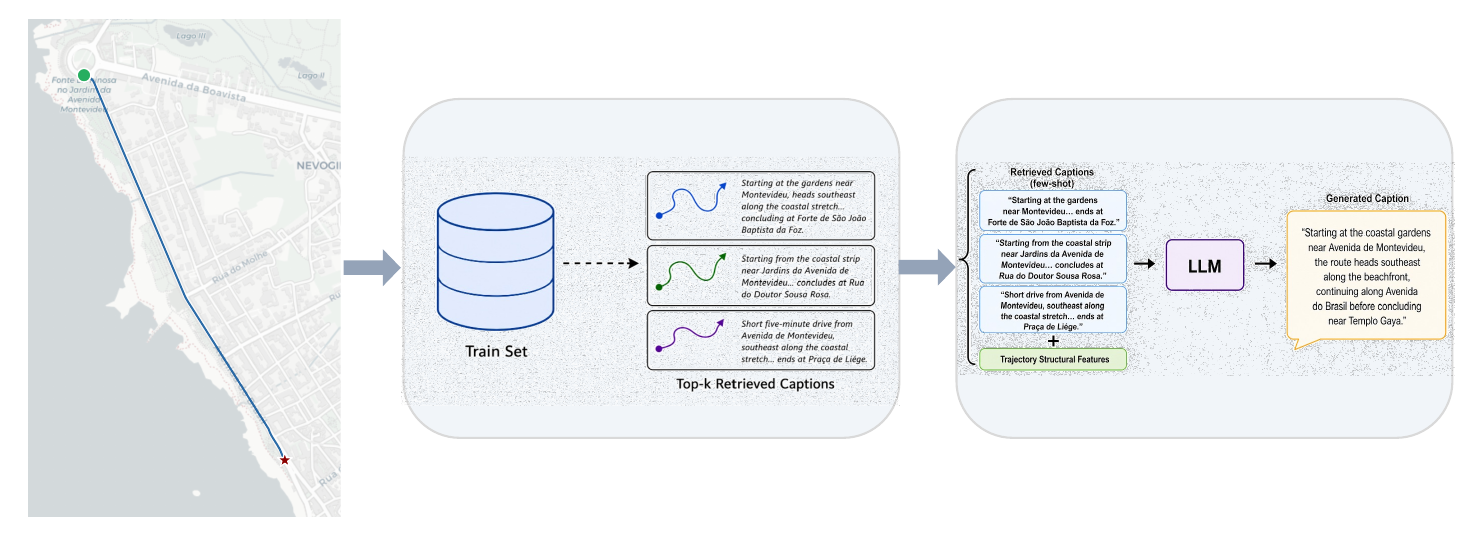}
    \caption{\textbf{TrajRap pipeline (Task~3).}
    Similar training trajectories are retrieved and their gold captions
    serve as few-shot examples, which are fed alongside the test
    trajectory's structural features to an LLM for factual captioning.}
    \label{fig:trajrap-pipeline}
\end{figure}

\textbf{TrajRap} (\textbf{R}etrieval-\textbf{A}ugmented \textbf{P}rofiling,
Task~3) generates factual trajectory captions through a
retrieval-augmented few-shot prompting pipeline
(Figure~\ref{fig:trajrap-pipeline}).
\textit{Step~1 (Example Retrieval):} given a test trajectory, we
encode its textual representation with a text embedding
model~\cite{nussbaum2024nomic} and retrieve the top-$K$ most
similar training trajectories via cosine similarity over a
pre-built index of the training set.
The gold captions of the retrieved trajectories are collected as
few-shot references.
\textit{Step~2 (Prompt Assembly):} the retrieved captions are
placed into the prompt as style and granularity references,
together with the test trajectory's structural features (bearing
changes, duration, road names) and, when available, H3-cell
semantic descriptions.
\textit{Step~3 (Caption Generation):} the assembled prompt is fed
to an LLM (Qwen-3.5 4B), which generates a grounded caption
conditioned on both the few-shot examples and the trajectory
context, without any model fine-tuning.
We additionally evaluate two zero-shot ablation baselines:
(i)~\textit{Struct.}: structural features only, no retrieval;
(ii)~\textit{Sem.}: structural + H3 semantic context, no
few-shot retrieval;
as well as a distillation variant,
(iii)~\textit{Sem.\ Distill.}: the LLM is LoRA
fine-tuned~\cite{hu2022lora} with full semantic context but tested
with structural features only, probing whether spatial semantics
can be internalized through training.

\begin{figure*}[htbp]
  \centering
  \includegraphics[width=\linewidth]{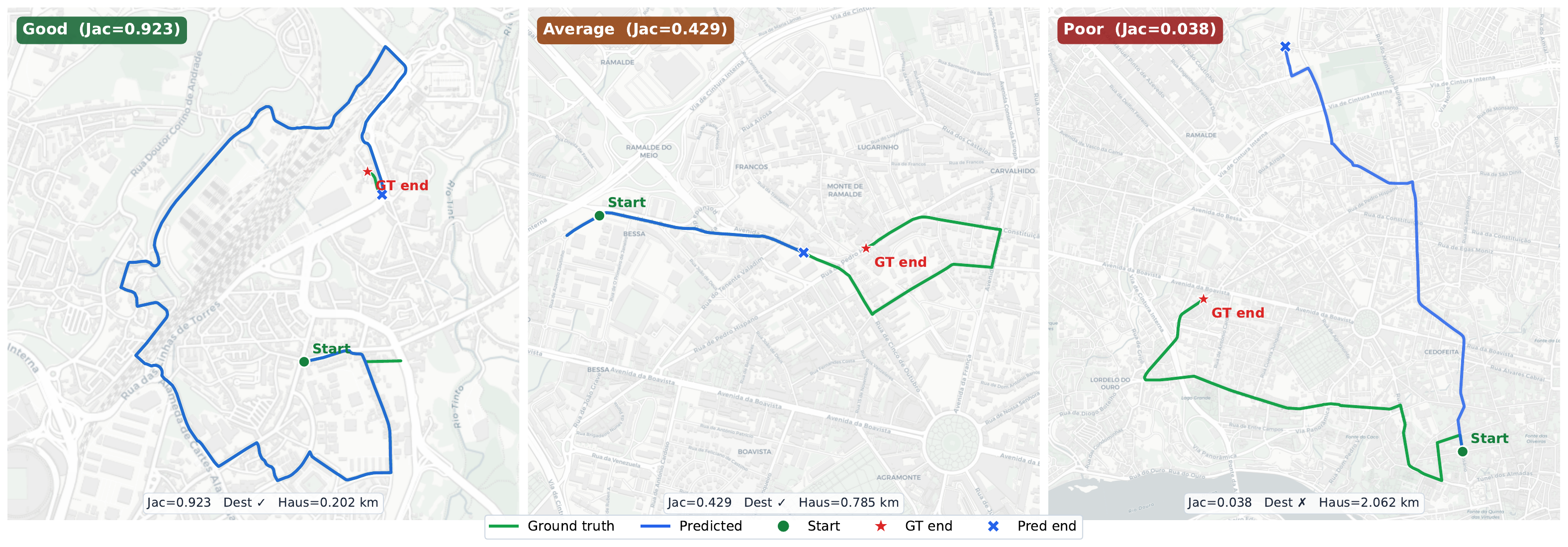}
  \caption{\textbf{Task~1 qualitative comparison: good, moderate, and poor predictions.}
  Each column shows the ground-truth trajectory (blue) and the TrajAnchor prediction (red)
  for one Porto test case. Left: the predicted route closely follows the ground truth
  with correct destination and waypoints. Middle: the destination is approximately correct
  but the predicted route diverges in the middle segment. Right: the destination is
  entirely wrong, producing a route with low spatial overlap.}
  \label{fig:task1-trio}
\end{figure*}

Figure~\ref{fig:task1-trio} presents three representative Task~1
predictions spanning a range of quality levels.
In the good case (left), TrajAnchor correctly identifies the
destination and reproduces the intended route shape, yielding
high Jaccard and low DTW.
The moderate case (middle) shows that even when the endpoint is
approximately correct, mid-route deviations (e.g., taking a
parallel street or missing a waypoint turn) substantially reduce
Jaccard while keeping endpoint distance low, highlighting the gap
between destination accuracy and full trajectory fidelity.
In the poor case (right), incorrect destination retrieval cascades
into a completely different route, confirming the oracle analysis
(Table~\ref{tab:task1_oracle}) that constraints selection is the
primary bottleneck in the current pipeline.

\begin{table*}[t]
  \centering
  \small
  \setlength{\tabcolsep}{4pt}
  \caption{Task 1 supplementary metrics not reported in Table~\ref{tab:task1_baselines}.
  H@$K$: destination hit within $K$ hops ($\uparrow$);
  Over-Rt.\ / Under-Rt.: fraction of routes $>$1.5$\times$ or $<$0.5$\times$ the ground-truth length ($\downarrow$).
  Best per dataset in \textbf{bold}, second best \underline{underlined}.
  $^\dagger$: Qwen3.5-4B (default); $^\star$: Claude Sonnet~4.6.}
  \label{tab:task1_supp}
  \begin{tabular}{llccccc}
  \toprule
  \toprule
  \textbf{Dataset} & \textbf{Method}
    & \textbf{H@1}$\uparrow$
    & \textbf{H@3}$\uparrow$
    & \textbf{H@10}$\uparrow$
    & \textbf{Over-Rt.}$\downarrow$
    & \textbf{Under-Rt.}$\downarrow$ \\
  \midrule
  \midrule
  \multirow{7}{*}{Porto}
    & DestSP (BM25)      & \textbf{0.0010} & 0.0527 & 0.1463 & 0.1869 & \textbf{0.2208} \\
    & DestSP (Embed)     & 0.0005 & 0.0618 & 0.1720 & 0.1338 & \underline{0.2506} \\
    \addlinespace[1.5pt]
    & ConstrSP           & 0.0005 & 0.0593 & 0.1658 & 0.2051 & 0.3532 \\
    \addlinespace[1.5pt]
    & TrajAnchor (Q3.5-2B) & 0.0008 & 0.0631 & 0.1918 & 0.0869 & 0.3455 \\
    & \textbf{TrajAnchor}$^\dagger$ & \underline{0.0009} & 0.0610 & 0.1977 & \underline{0.0656} & 0.3741 \\
    & \textbf{TrajAnchor}$^\star$ (CS4.6) & \textbf{0.0010} & \textbf{0.0694} & \textbf{0.2110} & 0.0758 & 0.3411 \\
    & TrajAnchor (Q3.5-9B) & 0.0005 & \underline{0.0634} & \underline{0.2050} & \textbf{0.0606} & 0.3729 \\
  \midrule
  \multirow{7}{*}{SF}
    & DestSP (BM25)      & 0.0002 & 0.0816 & 0.1961 & 0.2073 & \textbf{0.1704} \\
    & DestSP (Embed)     & 0.0003 & 0.0942 & 0.2380 & 0.1812 & \underline{0.1898} \\
    \addlinespace[1.5pt]
    & ConstrSP           & 0.0003 & 0.0913 & 0.2339 & 0.1458 & 0.4756 \\
    \addlinespace[1.5pt]
    & TrajAnchor (Q3.5-2B) & 0.0003 & 0.0990 & 0.2555 & 0.1457 & 0.2378 \\
    & \textbf{TrajAnchor}$^\dagger$ & \textbf{0.0004} & \underline{0.0999} & \underline{0.2589} & \textbf{0.1148} & 0.2760 \\
    & \textbf{TrajAnchor}$^\star$ (CS4.6) & \textbf{0.0004} & \textbf{0.1048} & \textbf{0.2740} & \underline{0.1256} & 0.2449 \\
    & TrajAnchor (Q3.5-9B) & 0.0003 & 0.0998 & 0.2552 & 0.1353 & 0.2434 \\
  \midrule
  \multirow{7}{*}{BJ}
    & DestSP (BM25)      & 0.0013 & 0.0699 & 0.1455 & 0.2869 & \textbf{0.1554} \\
    & DestSP (Embed)     & \underline{0.0022} & 0.1156 & 0.2224 & 0.2278 & \underline{0.1790} \\
    \addlinespace[1.5pt]
    & ConstrSP           & 0.0019 & 0.1084 & \underline{0.2575} & \textbf{0.1730} & 0.4483 \\
    \addlinespace[1.5pt]
    & TrajAnchor (Q3.5-2B) & \textbf{0.0024} & 0.1194 & 0.2323 & 0.2097 & 0.2041 \\
    & \textbf{TrajAnchor}$^\dagger$ & \underline{0.0022} & 0.1250 & 0.2391 & 0.2034 & 0.2005 \\
    & \textbf{TrajAnchor}$^\star$ (CS4.6) & \underline{0.0022} & \textbf{0.1400} & \textbf{0.2586} & \underline{0.1932} & 0.1920 \\
    & TrajAnchor (Q3.5-9B) & 0.0020 & \underline{0.1297} & 0.2434 & 0.2030 & 0.1963 \\
  \bottomrule
  \bottomrule
  \end{tabular}
\end{table*}

Table~\ref{tab:task1_supp} provides additional destination-proximity
and routing diagnostics.
H@1 is near zero for all methods, confirming that exact one-hop
destination matching is extremely rare, but H@10 reaches
0.20--0.27, indicating that predicted endpoints fall within the
correct neighborhood.
TrajAnchor variants achieve the lowest Over-Routing rates but
exhibit elevated Under-Routing, indicating a tendency to generate
shorter-than-intended routes.
DestSP baselines show more balanced Over/Under-Routing, as
shortest-path routing naturally calibrates route length when the
destination is approximately correct.
ConstrSP shows the highest Under-Routing across all cities,
suggesting that added waypoint constraints sometimes cause
premature route termination.
\begin{table*}[t]
  \centering
  \footnotesize
  \setlength{\tabcolsep}{4pt}
  \caption{Task 1 oracle analysis: performance upper bound when the best destination candidate is selected from the retrieval pool.
  Metrics mirror Table~\ref{tab:task1_baselines} (O- prefix = oracle counterpart).
  ConstrSP is omitted as it does not use retrieval.
  Best per dataset in \textbf{bold}, second best \underline{underlined}.
  $^\dagger$: Qwen3.5-4B (default); $^\star$: Claude Sonnet~4.6.}
  \label{tab:task1_oracle}
  \begin{tabular}{llccccccc}
  \toprule
  \toprule
  \textbf{Dataset} & \textbf{Method}
    & \textbf{O-H@5}$\uparrow$
    & \textbf{O-Dest}$\uparrow$
    & \textbf{O-Dist(km)}$\downarrow$
    & \textbf{O-Jac}$\uparrow$
    & \textbf{O-DTW}$\downarrow$
    & \textbf{O-Haus}$\downarrow$
    & \textbf{O-EDR}$\downarrow$ \\
  \midrule
  \midrule
  \multirow{6}{*}{Porto}
    & DestSP (BM25)      & 0.1929 & 0.2019 & 1.8152 & 0.3816 & 40.2706 & 1.6284 & 0.6546 \\
    & DestSP (Embed)     & 0.2758 & 0.2948 & 1.3101 & 0.4622 & 29.9528 & 1.2854 & 0.5830 \\
    \addlinespace[1.5pt]
    & TrajAnchor (Q3.5-2B) & 0.3926 & 0.4096 & 1.0963 & 0.5298 & 26.4659 & 1.1205 & 0.5222 \\
    & \textbf{TrajAnchor}$^\dagger$ & \underline{0.4398} & \underline{0.4612} & \underline{0.9405} & \underline{0.5671} & \underline{23.3307} & \underline{1.0069} & \underline{0.4893} \\
    & \textbf{TrajAnchor}$^\star$ (CS4.6) & 0.4389 & 0.4592 & 0.9421 & 0.5628 & 23.3345 & 1.0089 & 0.4931 \\
    & TrajAnchor (Q3.5-9B) & \textbf{0.4497} & \textbf{0.4706} & \textbf{0.9170} & \textbf{0.5726} & \textbf{22.7166} & \textbf{0.9906} & \textbf{0.4845} \\
  \midrule
  \multirow{6}{*}{SF}
    & DestSP (BM25)      & 0.2049 & 0.1916 & 1.9974 & 0.3227 & 39.3191 & 1.7641 & 0.7255 \\
    & DestSP (Embed)     & 0.2746 & 0.2596 & 1.6100 & 0.3744 & 32.1121 & 1.5190 & 0.6814 \\
    \addlinespace[1.5pt]
    & TrajAnchor (Q3.5-2B) & 0.3216 & 0.3037 & 1.4080 & 0.4093 & 28.4060 & 1.3716 & 0.6514 \\
    & \textbf{TrajAnchor}$^\dagger$ & \underline{0.3563} & \underline{0.3382} & \textbf{1.2413} & \textbf{0.4400} & \textbf{25.1852} & \textbf{1.2462} & \textbf{0.6245} \\
    & \textbf{TrajAnchor}$^\star$ (CS4.6) & \textbf{0.3648} & \textbf{0.3467} & \underline{1.2765} & \underline{0.4389} & \underline{25.9969} & \underline{1.2725} & \underline{0.6247} \\
    & TrajAnchor (Q3.5-9B) & 0.3413 & 0.3221 & 1.2966 & 0.4278 & 26.2175 & 1.2892 & 0.6340 \\
  \midrule
  \multirow{6}{*}{BJ}
    & DestSP (BM25)      & 0.1563 & 0.1378 & 3.3510 & 0.3124 & 56.0179 & 2.9876 & 0.6871 \\
    & DestSP (Embed)     & 0.2621 & 0.2373 & 2.6645 & 0.3920 & 44.5428 & 2.4794 & 0.6114 \\
    \addlinespace[1.5pt]
    & TrajAnchor (Q3.5-2B) & 0.2770 & 0.2501 & 2.5339 & 0.4072 & 42.2292 & 2.3745 & 0.5971 \\
    & \textbf{TrajAnchor}$^\dagger$ & 0.2819 & 0.2550 & 2.4807 & 0.4128 & 41.2324 & 2.3370 & 0.5905 \\
    & \textbf{TrajAnchor}$^\star$ (CS4.6) & \textbf{0.3138} & \textbf{0.2850} & \textbf{2.3683} & \textbf{0.4299} & \textbf{39.6896} & \textbf{2.2488} & \textbf{0.5749} \\
    & TrajAnchor (Q3.5-9B) & \underline{0.2881} & \underline{0.2606} & \underline{2.4688} & \underline{0.4154} & \underline{41.0469} & \underline{2.3258} & \underline{0.5888} \\
  \bottomrule
  \bottomrule
  \end{tabular}
\end{table*}

Table~\ref{tab:task1_oracle} reveals the performance ceiling
under oracle destination selection.
All metrics improve substantially: for example, Porto Jaccard
rises from 0.25 (Table~\ref{tab:task1_baselines}) to 0.57 under
oracle, and endpoint distance drops by over 40\%.
This gap confirms that destination selection, rather than the
routing algorithm itself, is the primary bottleneck in
the current pipeline.
Larger LLM backbones consistently improve oracle performance,
indicating that stronger language-to-map grounding yields
higher-quality destination candidates even when the final
selection is imperfect.
Beijing retains the largest absolute DTW and Hausdorff values
even under oracle, reflecting the intrinsic difficulty of its
longer and more complex road network.

\begin{table*}[t]
  \centering
  \small
  \setlength{\tabcolsep}{4pt}
  \caption{Task 2 supplementary retrieval metrics (J@10, SR@10, R@50) across all cities.
  Combined with Table~\ref{tab:task2_retrieval} for full results. Best per dataset in \textbf{bold}, second best \underline{underlined}.}
  \label{tab:task2_supp}
  \begin{tabular}{llccc}
  \toprule
  \toprule
  \textbf{Dataset} & \textbf{Method} & \textbf{J@10} & \textbf{SR@10} & \textbf{R@50} \\
  \midrule
  \midrule
  \multirow{10}{*}{Porto}
    & TF-IDF + SVD          & 0.5529 & 0.2696 & 0.2837 \\
    & BoW + SVD             & 0.5520 & 0.2681 & 0.2858 \\
  \cmidrule(lr){2-5}
    & UniTraj               & 0.4151 & 0.1418 & 0.1880 \\
    & TrajCL                & 0.6189 & 0.4132 & 0.4804 \\
    & T-JEPA                & 0.5759 & 0.3498 & 0.4128 \\
    & BLUE                  & 0.5113 & 0.2544 & 0.3000 \\
  \cmidrule(lr){2-5}
    &    Sem. (Zero-shot)   & 0.2786 & 0.1320 & 0.2140 \\
    &    Sem.               & \underline{0.6304} & \textbf{0.4425} & \textbf{0.5491} \\
    & Road GATv2            & 0.5492 & 0.2793 & 0.3256 \\
    & \textbf{TrajFuse}     & \textbf{0.6426} & \underline{0.4401} & \underline{0.5286} \\
  \midrule\midrule
  \multirow{10}{*}{SF}
    & TF-IDF + SVD          & 0.4230 & 0.1531 & 0.2472 \\
    & BoW + SVD             & 0.4199 & 0.1482 & 0.2361 \\
  \cmidrule(lr){2-5}
    & UniTraj               & 0.1978 & 0.0534 & 0.0852 \\
    & TrajCL                & \underline{0.5951} & 0.4680 & 0.5965 \\
    & T-JEPA                & 0.4328 & 0.2682 & 0.3995 \\
    & BLUE                  & 0.4152 & 0.2444 & 0.3822 \\
  \cmidrule(lr){2-5}
    &    Sem. (Zero-shot)   & 0.2805 & 0.1887 & 0.2807 \\
    &    Sem.               & 0.5892 & \underline{0.4730} & \textbf{0.6548} \\
    & Road GATv2            & 0.4286 & 0.2021 & 0.3395 \\
    & \textbf{TrajFuse}     & \textbf{0.6304} & \textbf{0.5082} & \underline{0.6372} \\
  \midrule\midrule
  \multirow{10}{*}{BJ}
    & TF-IDF + SVD          & 0.3296 & 0.1080 & 0.2342 \\
    & BoW + SVD             & 0.3218 & 0.1024 & 0.2246 \\
  \cmidrule(lr){2-5}
    & UniTraj               & 0.0730 & 0.0255 & 0.0534 \\
    & TrajCL                & \underline{0.5318} & 0.4219 & 0.6034 \\
    & T-JEPA                & 0.4489 & 0.3320 & 0.5105 \\
    & BLUE                  & 0.3009 & 0.1440 & 0.3072 \\
  \cmidrule(lr){2-5}
    &    Sem. (Zero-shot)   & 0.3056 & 0.2575 & 0.3733 \\
    &    Sem.               & 0.5280 & \underline{0.4443} & \underline{0.6248} \\
    & Road GATv2            & 0.3957 & 0.2246 & 0.4046 \\
    & \textbf{TrajFuse}     & \textbf{0.5969} & \textbf{0.5021} & \textbf{0.6835} \\
  \bottomrule
  \bottomrule
  \end{tabular}
  \vspace{-8pt}
  \end{table*}

Table~\ref{tab:task2_supp} extends the retrieval evaluation with
J@10, SR@10, and R@50.
TrajFuse leads on J@10 across all three cities
(Porto 0.64, SF 0.63, BJ 0.60), confirming that fusing geometric
and semantic representations yields the best spatial overlap at
deeper retrieval depths.
The fine-tuned Sem.\ encoder is competitive on SR@10 and R@50,
occasionally surpassing TrajFuse (e.g., Porto R@50: 0.55 vs.\ 0.53),
suggesting that pure text-aligned embeddings excel at retrieving
at least one high-quality match.
Among geometric-only baselines, TrajCL consistently outperforms
UniTraj, T-JEPA, and BLUE, while UniTraj collapses on BJ
(J@10 = 0.07), highlighting the difficulty of cross-city
generalization for pre-trained trajectory encoders.
Zero-shot Sem.\ trails all fine-tuned methods, confirming that
contrastive fine-tuning on in-domain trajectory--text pairs is
essential for competitive retrieval.

\begin{figure*}[htbp]
  \centering
  \includegraphics[width=\linewidth]{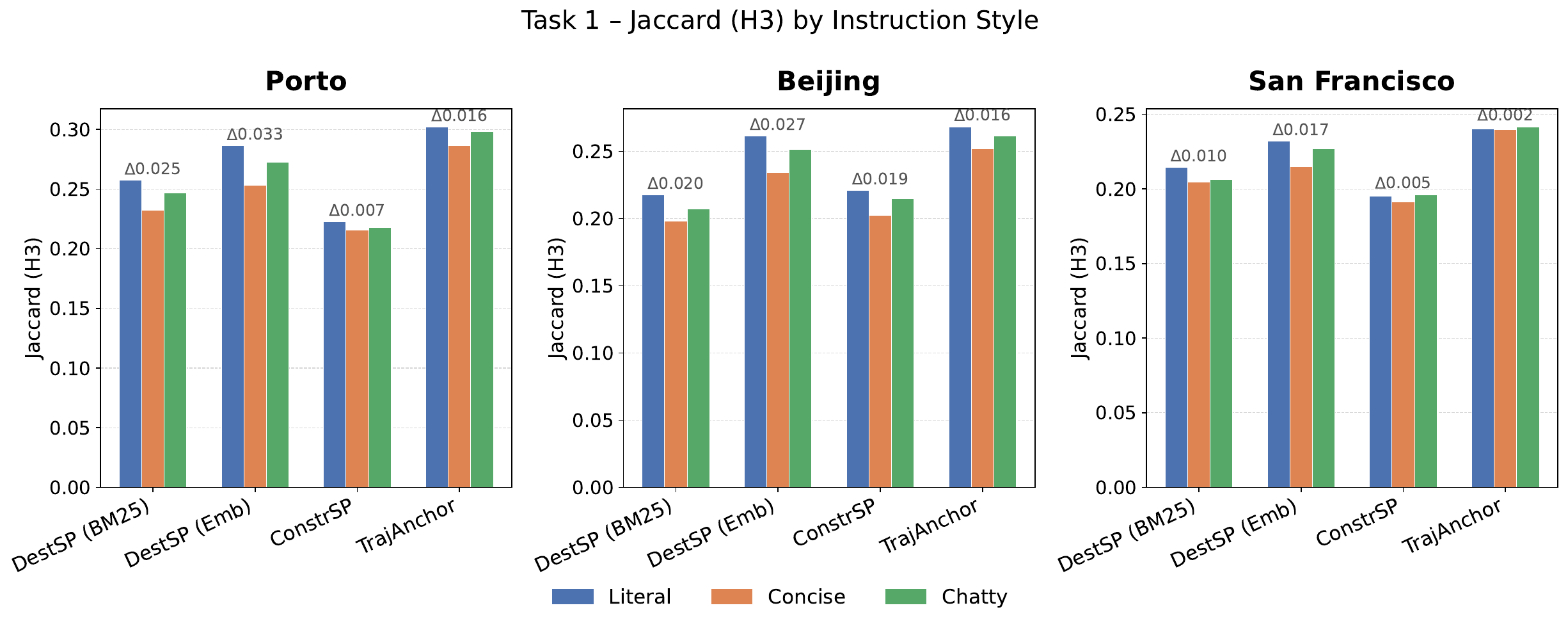}
  \caption{\textbf{Task~1 Jaccard (H3) by instruction style} across three cities.
  Performance is broken down by the three instruction variants
  (Literal, Concise, Chatty) for each baseline and TrajAnchor.}
  \label{fig:task1-style-bars}
\end{figure*}

Figure~\ref{fig:task1-style-bars} compares Task~1 trajectory generation
quality (Jaccard over H3 cells) across the three instruction styles.
Literal instructions consistently yield the highest fidelity,
as they provide explicit route constraints,
while Concise and Chatty variants degrade gracefully.
TrajAnchor maintains relatively stable performance across styles
compared to the DestSP baselines,
suggesting that retrieval-augmented routing is more robust to
variations in instruction verbosity.

\begin{figure*}[htbp]
  \centering
  \includegraphics[width=\linewidth]{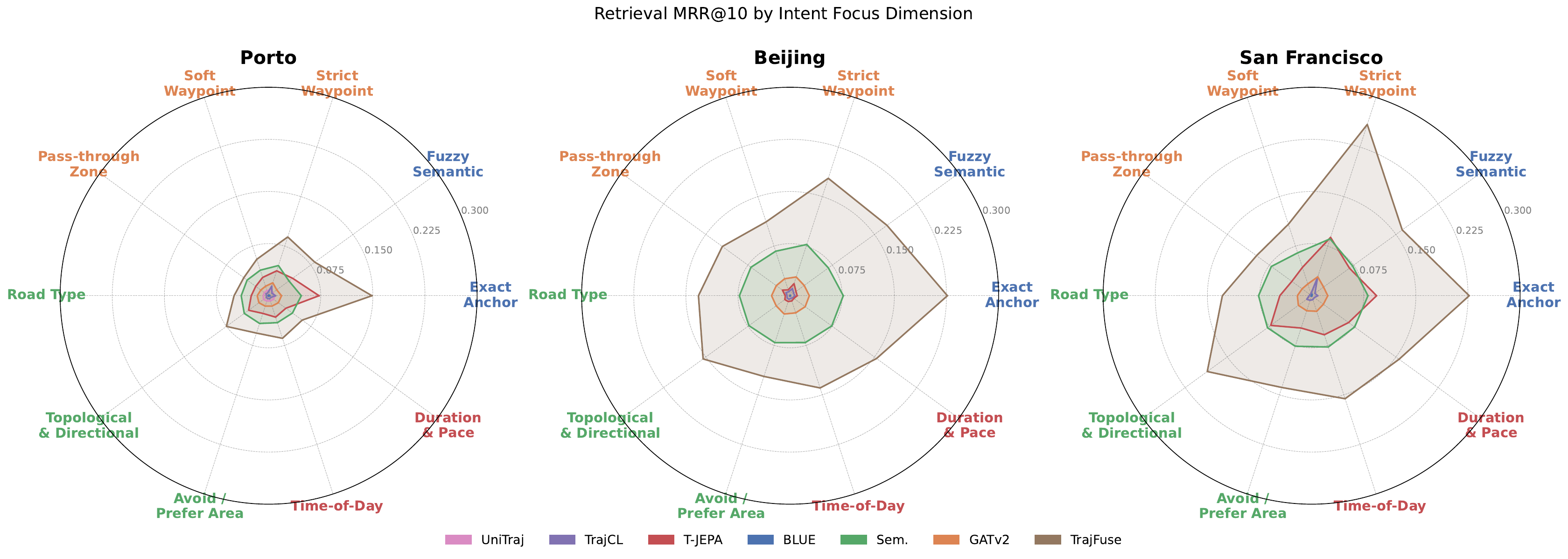}
  \caption{\textbf{Task~2 retrieval MRR@10 by intent focus dimension} across
  three cities. Each axis corresponds to one of the ten intent subcategories
  in the TrajPrism taxonomy.}
  \label{fig:retrieval-radar}
  \vspace{-8pt}
\end{figure*}

Figure~\ref{fig:retrieval-radar} breaks down Task~2 retrieval performance
(MRR@10) by intent focus dimension.
TrajFuse and the fine-tuned Sem.\ encoder dominate on most axes,
but performance varies sharply across intent types:
\emph{Exact Anchor} and \emph{Strict Waypoint} queries (where the target is spatially unambiguous) yield the highest MRR across all
methods, while \emph{Duration \& Pace} and \emph{Avoid/Prefer Area}
remain consistently difficult, suggesting that temporal and
preference-based intents are harder to capture in current
trajectory--text alignment.
Purely geometric encoders (UniTraj, BLUE) trail on nearly every
dimension, confirming that semantic augmentation is critical for
language-driven retrieval.
Beijing exhibits the steepest cross-dimension variance, reflecting
the greater spatial complexity of its road network.

\subsection{TrajPrism Construction Pipeline Details}

\subsubsection{Trajectory and Map Inputs}\label{sec:input}

Table~\ref{tab:input_fields} summarizes the raw data fields consumed by
the TrajPrism pipeline, organized by source category.
Trajectory records provide map-matched road segment sequences with
timestamps; per-segment attributes (name, bearing, class) are
retrieved from OpenStreetMap; and per-H3-cell semantic context is
derived from Overture Maps POI/land-use annotations augmented by
a GNN-based geo-tag classifier and an LLM-generated (Qwen3-4B-Instruct) area narrative.

\begin{table*}[htbp]
  \centering
  \footnotesize
  \setlength{\tabcolsep}{5pt}
  \caption{Input data sources and per-record fields used in the TrajPrism pipeline.}
  \label{tab:input_fields}
  \begin{tabular}{llll}
  \toprule
  \textbf{Category} & \textbf{Field} & \textbf{Source} & \textbf{Description} \\
  \midrule
  \multirow{3}{*}{Trajectory}
    & \texttt{rid\_list}   & Map matching   & Ordered road segment IDs \\
    & \texttt{time\_list}  & GPS data       & Per-segment timestamps \\
    & \texttt{mm\_id}      & Map matching   & Map-matched trajectory identifier \\
  \midrule
  \multirow{4}{*}{Road Segment}
    & Road name     & OSM            & Human-readable street name \\
    & Bearing       & Computed       & Segment direction as $(\sin\theta,\cos\theta)$ \\
    & Length        & OSM geometry   & Polyline length (m) \\
    & Road class    & OSM            & \texttt{highway} tag (20 classes) \\
  \midrule
  \multirow{6}{*}{H3 Cell}
    & Geo-tag       & GNN classifier & Urban/Inland, Waterfront, Green/Park, Coastal/Beach \\
    & Natural env.  & Overture Maps  & Beach, forest, park, water, etc. \\
    & Commercial    & Overture Maps  & Dining, entertainment, services, shopping \\
    & Public fac.   & Overture Maps  & Education, government, healthcare, transport \\
    & Residential   & Overture Maps  & High-density, low-density, mixed \\
    & Named POIs    & Overture Maps  & Place names and POI categories \\
  \midrule
  \multirow{3}{*}{H3 Road Net.}
    & Class lengths & Computed       & Road length (m) per class within the cell \\
    & Neighbors     & H3 library     & Adjacent H3 cell IDs (resolution 9) \\
    & Narrative     & LLM-generated  & Natural-language area description \\
  \bottomrule
  \end{tabular}
  \vspace{-16pt}
  \end{table*}

\subsubsection{Phase Compression}\label{sec:phase-compression}

\begin{figure}[htbp]
  \centering
  \includegraphics[width=0.75\linewidth]{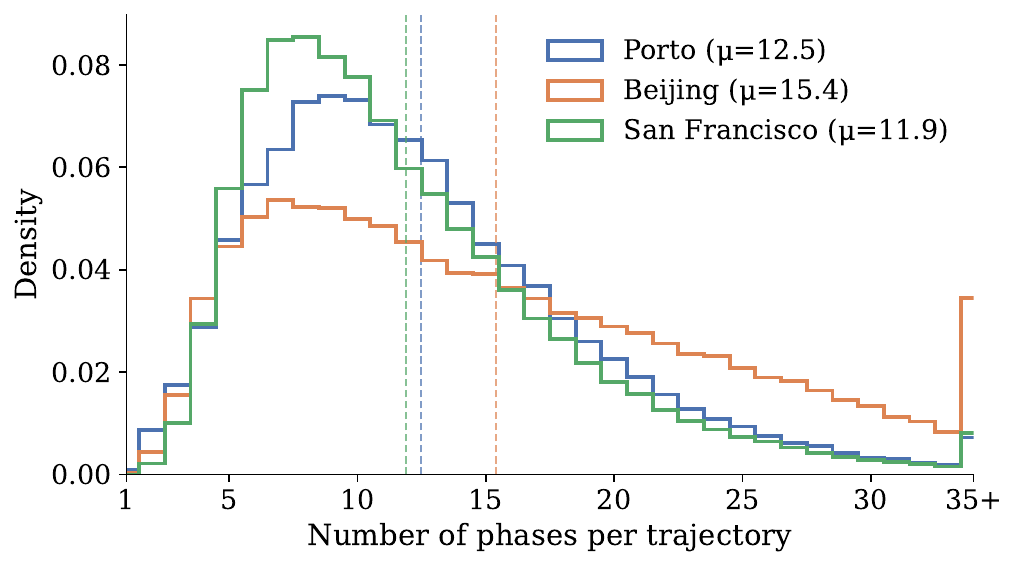}
  \caption{Distribution of the number of phases per trajectory after H3-based
  compression. The average trajectory is reduced to 12--15 phases across all
  three cities, with Beijing exhibiting a higher mean due to longer average
  trip lengths.}
  \label{fig:phase_count_dist}
\end{figure}

Each phase is represented as a tuple:
\begin{equation}
    \mathbf{p}_k = (h_k,\ n_k,\ d_k,\ \Delta t_k,\ \rho_k,\ \mathbf{m}_k,\ s_k)
\end{equation}
Here $h_k$ is the dominant H3 cell of phase $P_k$, $n_k=|P_k|$ is the segment 
count, and $d_k$ is the phase-level heading computed by aggregating segment 
bearings and discretizing the result into eight compass directions. 
The duration $\Delta t_k$ is derived from timestamps aligned with the first and 
last segments of the phase, while $\rho_k \in \{\text{O, T, D}\}$ denotes whether 
the phase is the Origin, Transit, or Destination according to its position in the 
trajectory. 
$\mathbf{m}_k$ is the set of deduplicated road names traversed in the phase, 
included when available.
Finally, $s_k$ is the semantic area description associated with the phase's H3 
cell, incorporating road-network structure and POI context from the map-derived 
annotations introduced above. 
Figure~\ref{fig:phase_count_dist} shows that this compression typically reduces 
trajectories to 12--15 phases across all three cities.

Below is the compressed representation of a trajectory (Porto), 
reduced from 58 road segments to 16 semantic phases over 8 min 15 sec. 
We show the Origin, one representative Transit phase, and the Destination.

\vspace{-6pt}

\noindent
\begin{minipage}{\linewidth}
\begin{lstlisting}[language=json, basicstyle=\ttfamily\footnotesize,
  frame=single, breaklines=true, columns=fullflexible,
  showstringspaces=false,
  caption={Compressed phase sequence for an Porto trajectory.}]
{
  "traj_id": 1626278,
  "meta": {
    "n_rids": 58, "n_phases": 16,
    "start_time": "Saturday, Jun 14, 2014 at 4:11 AM",
    "total_duration": "8 min 15 sec"
  },
  "phases": [
    {
      "p": 0, "role": "O", "dir": "NW",
      "n": 4, "duration": "55 sec",
      "road_names": ["Rua Prof. Vicente Jose de Carvalho"],
      "desc": "GNN: WATERFRONT | Natural: garden, park, water | Commercial: dining, shopping | Facilities: bar, coffee shop | District: Cedofeita"
    },
    {
      "p": 1, "role": "T", "dir": "SW",
      "n": 2, "duration": "15 sec",
      "road_names": ["Rua de Clemente Meneres"],
      "desc": "GNN: COASTAL/BEACH | Natural: park, rock, tree | Commercial: dining | Facilities: art gallery, restaurant | District: Cedofeita"
    },
    // ... phases 2--14 (Transit) ...
    {
      "p": 15, "role": "D", "dir": "NE",
      "n": 1, "duration": "0 sec",
      "road_names": ["Rua Direita das Campinas"],
      "desc": "GNN: GREEN/PARK | Natural: tree | Commercial: shopping | Facilities: supermarket | District: Ramalde"
    }
  ]
}
\end{lstlisting}
\end{minipage}

\subsubsection{Intent Sampling}\label{sec:intent-sampling}

\begin{figure}[htbp]
  \centering
  \includegraphics[width=\linewidth]{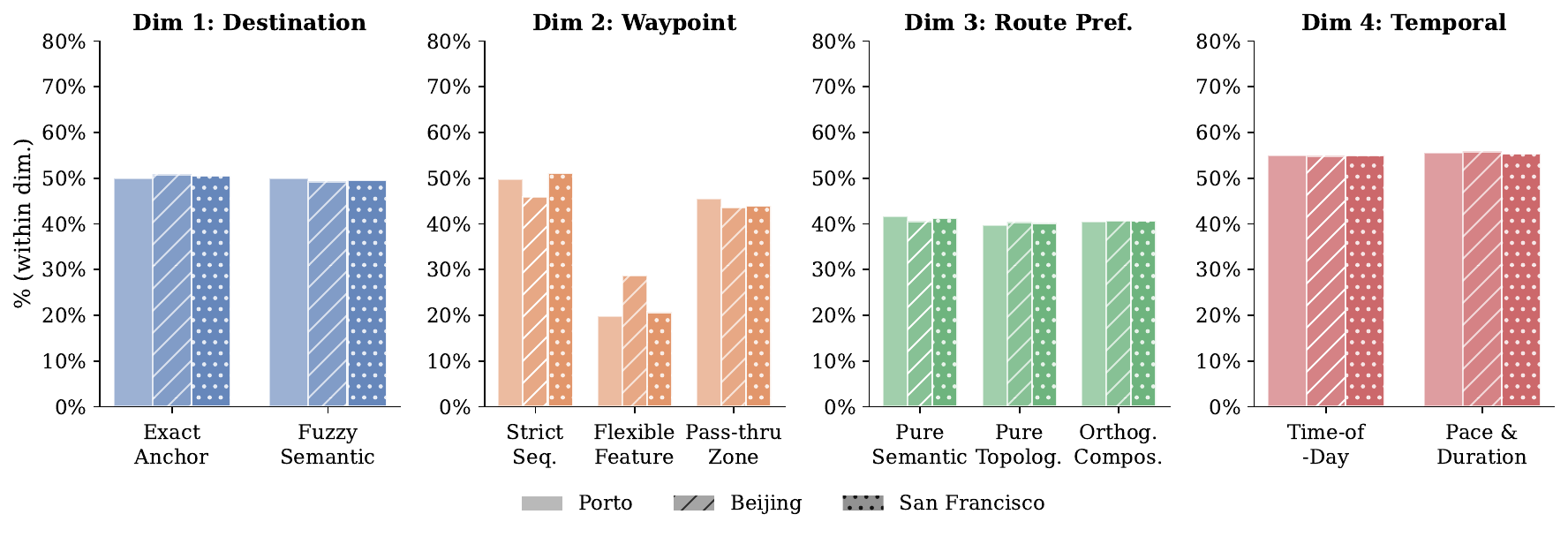}
  \caption{Conditional distribution of fine-grained semantic subcategories in
  TrajPrism. Percentages are normalized within each semantic dimension, showing
  that the benchmark evenly covers diverse destination references, waypoint constraints,
  route preferences, and temporal or pace cues rather than relying on a single
  instruction pattern.}
  \label{fig:dataset_subcategory_conditional}
\end{figure}

Table~\ref{tab:intent_taxonomy} lists the full four-dimensional
travel-intent taxonomy used to construct structured intent profiles.
Figure~\ref{fig:dataset_subcategory_conditional} shows the conditional distribution of fine-grained scenarios within each dimension, 
confirming balanced coverage across all 10 subcategories.
Each trajectory is assigned $k$ scenarios sampled without replacement,
where $k \in \{1,2,3,4,5\}$ follows a weighted distribution
($p = [0.15, 0.35, 0.30, 0.15, 0.05]$) centered at 2--3 scenarios.
Single-scenario intents ($k{=}1$) are restricted to Dimension~1
(Destination only), reflecting simple navigation requests such as
``take me to the hospital''.
Composite intents ($k \geq 2$) always include exactly one Destination
scenario and draw the remaining scenarios from Dimensions~2--4,
producing realistic multi-constraint requests such as
``drive to the station, stop for fuel on the way, and avoid the highway''.
Figure~\ref{fig:scenario_count_dist} shows the resulting distribution
of scenario counts across the dataset, confirming that the majority of
trajectories carry 2--3 intent dimensions while preserving a meaningful
tail of simpler and more complex requests.

\begin{table}[t]
  \centering
  \small
  \setlength{\tabcolsep}{4pt}
  \caption{Four-dimensional travel-intent taxonomy used for intent sampling.
  Each trajectory is assigned 1--5 scenarios sampled across these dimensions.}
  \label{tab:intent_taxonomy}
  \begin{tabular}{@{}llp{0.52\linewidth}@{}}
  \toprule
  \textbf{Dimension} & \textbf{Scenario} & \textbf{Description} \\
  \midrule
  \multirow{2}{*}{1.\ Destination}
    & 1.1 Exact Anchor        & Specific physical destination (POI or road name) \\
    & 1.2 Fuzzy Semantic       & Conceptual destination (e.g.\ ``a quiet green area'') \\
  \midrule
  \multirow{3}{*}{2.\ Waypoint}
    & 2.1 Strict Sequential    & Must pass through specific named road segments \\
    & 2.2 Flexible / Feature   & Stop described by semantic features only \\
    & 2.3 Pass-through Zone    & Cross an area type without stopping \\
  \midrule
  \multirow{3}{*}{3.\ Route Pref.}
    & 3.1 Semantic Constraints  & Affinity or avoidance (e.g.\ parks, industrial zones) \\
    & 3.2 Topological / Direct. & Fluency, permeability, or directional preference \\
    & 3.3 Orthogonal Comp.      & Semantic vs.\ topology (e.g.\ ``through chaos but main road'') \\
  \midrule
  \multirow{2}{*}{4.\ Temporal/Pace}
    & 4.1 Time-of-Day          & Route choice driven by time or day of week \\
    & 4.2 Pace / Duration      & Urgency, leisure, or deadline-driven constraints \\
  \bottomrule
  \end{tabular}
\end{table}

\subsubsection{Persona and Generation Style Statistics}\label{sec:data_scenarios}

\begin{figure}[htbp]
  \centering
  \includegraphics[width=0.75\linewidth]{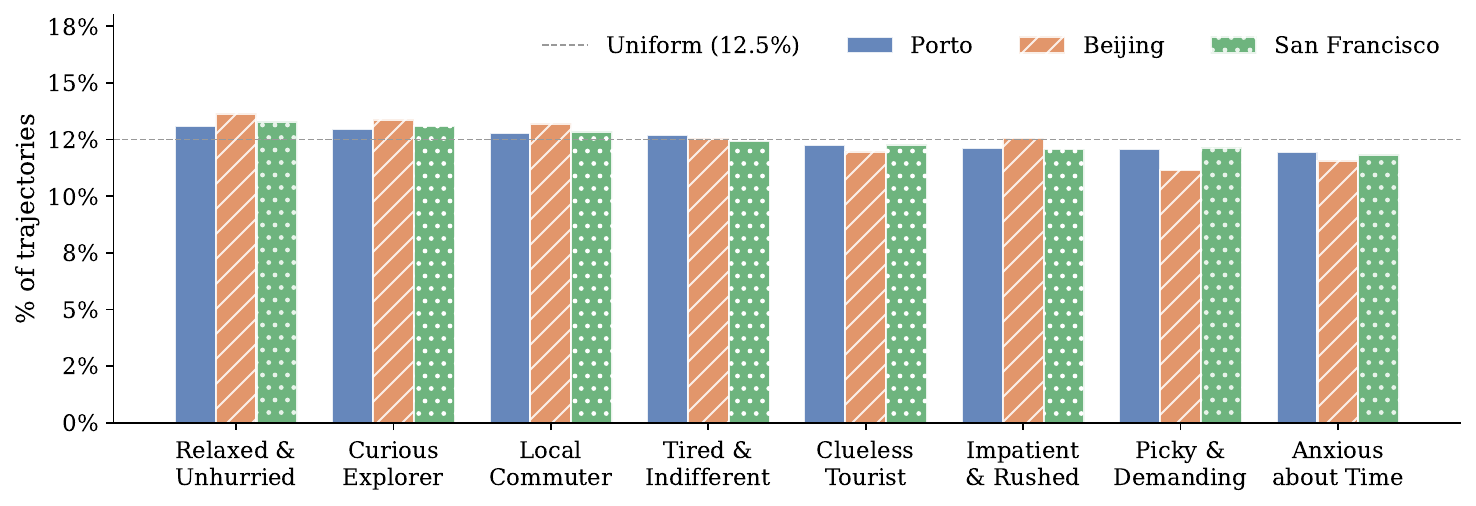}
  \caption{Distribution of personas used for generating navigation instructions in
  TrajPrism. Uniform persona sampling promotes balanced exposure to diverse tones,
  preferences, and communication styles, enabling models to learn language-grounded
  navigation behavior beyond a single dominant voice.} \looseness -1
  \label{fig:dataset_persona_distribution}
\end{figure}

\begin{figure}[htbp]
  \centering
  \includegraphics[width=0.75\linewidth]{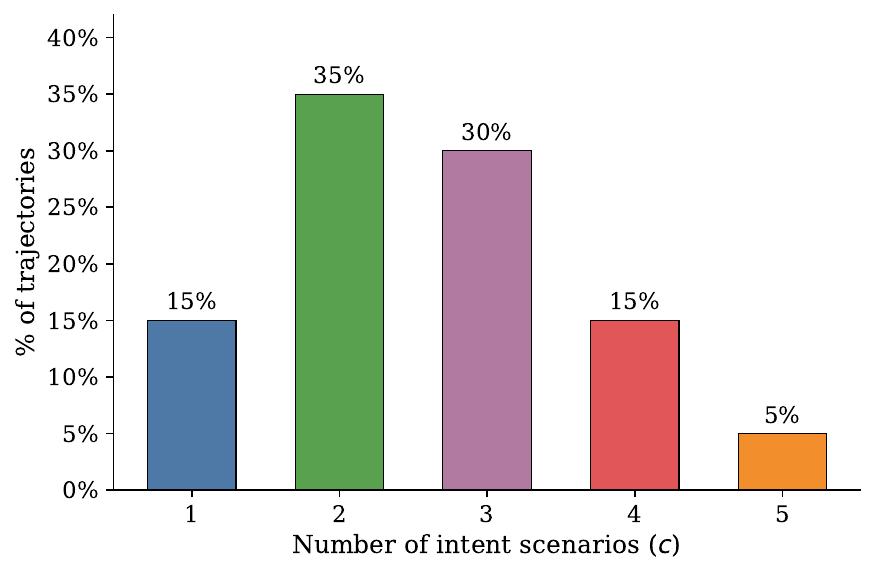}
  \caption{Distribution of the number of intent scenarios assigned to each
  generated trajectory. The sampling strategy mixes simple destination-only
  requests with multi-intent instructions that combine destination, waypoint,
  route-preference, and temporal or pace constraints.}
  \label{fig:scenario_count_dist}
\end{figure}

\begin{figure}[htbp]
  \centering
  \includegraphics[width=0.75\linewidth]{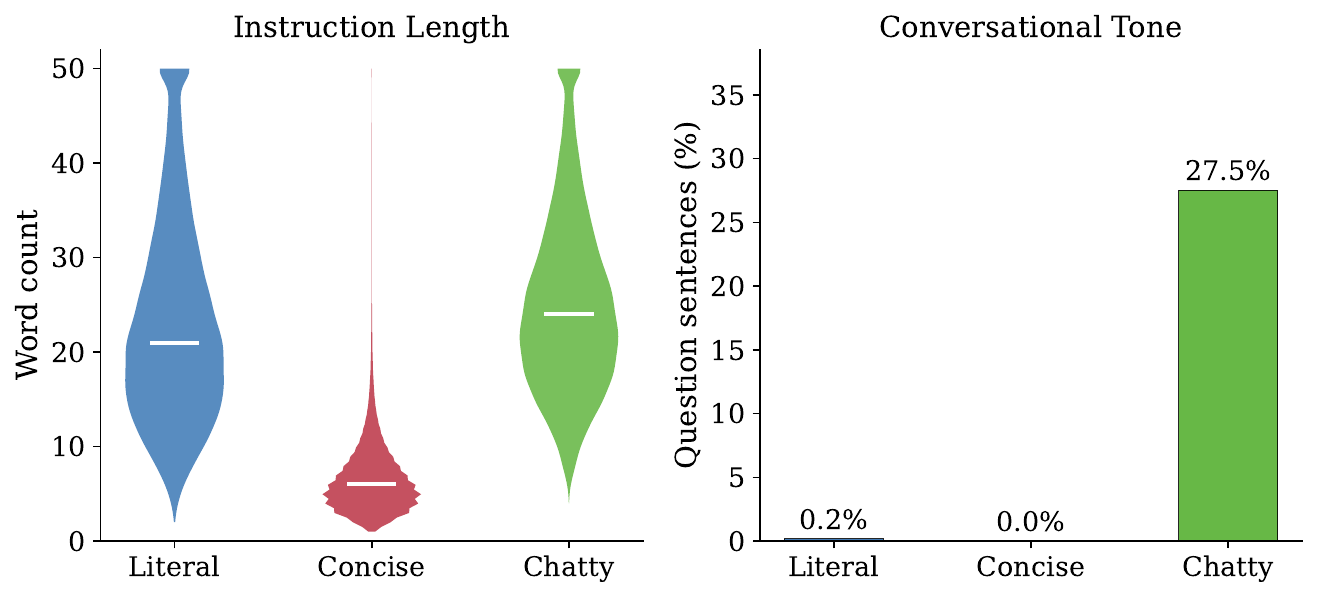}
  \caption{Token length distribution of generated navigation instructions
  across three stylistic variants (Literal, Concise, Chatty), confirming
  that the style and length guidance produces distinct verbosity profiles.}
  \label{fig:instruction_style_stats}
\end{figure}

To faithfully satisfy the combination of intent constraints, speaker persona, 
and style requirements, the model first produces a chain-of-thought analysis 
of the trajectory's semantic and topological features before generating 
task outputs for Tasks 1 and 2.
Figures~\ref{fig:dataset_persona_distribution}--\ref{fig:instruction_style_stats} and Table~\ref{tab:style_hints} summarize the resulting distributions of personas, 
scenario counts, instruction styles, and generation guidance.

\begin{table*}[htbp]
  \centering
  \caption{Style and length guidance hints sampled per generation. One sentence-form hint and one length hint are drawn uniformly at random per style per trajectory.}
  \label{tab:style_hints}
  \small
  \setlength{\tabcolsep}{5pt}
  \renewcommand{\arraystretch}{1.1}
  \begin{tabular}{lp{0.82\linewidth}}
  \toprule
  \textbf{Style} & \textbf{Sentence-form hint pool} \\
  \midrule
  Literal  & imperative command $\cdot$ declarative statement $\cdot$ sequenced action $\cdot$ context-aware continuation $\cdot$ telegraphic/terse $\cdot$ coordinate-style \\
  Concise  & single word or minimal fragment $\cdot$ terse command $\cdot$ abbreviated phrase $\cdot$ constraint-first $\cdot$ destination-only \\
  Chatty   & question form $\cdot$ complaint or reaction $\cdot$ narrative/storytelling $\cdot$ casual suggestion $\cdot$ soft request $\cdot$ trailing/open-ended \\
  \bottomrule
  \end{tabular}

  \vspace{6pt}

  \setlength{\tabcolsep}{5pt}
  \renewcommand{\arraystretch}{1.1}
  \begin{tabular}{ll}
  \toprule
  \textbf{Style} & \textbf{Length hint pool} \\
  \midrule
  \multirow{3}{*}{Literal}
    & Brief but complete (one sentence) \\
    & Moderate (dest.\ + 1--2 constraints) \\
    & Detailed (dest.\ + waypoints + constr.) \\
  \midrule
  \multirow{3}{*}{Concise}
    & Ultra-terse (telegram, fragments ok) \\
    & Short phrase (minimal thought) \\
    & Brief sentence (concise, grammatical) \\
  \midrule
  \multirow{3}{*}{Chatty}
    & Casual one-liner \\
    & Conversational (a couple sentences) \\
    & Chatty and detailed (rambling ok) \\
  \bottomrule
  \end{tabular}
\end{table*}

\subsubsection{Dataset Judgement}\label{sec:dataset-judging}

\begin{table}[htbp]
  \centering
  \caption{LLM and human evaluation rubric. Each criterion is scored on a 1--5 scale:
  5 = perfect/excellent, 4 = good with minor flaws, 3 = acceptable but limited,
  2 = poor with severe issues, and 1 = failed/completely wrong.}
  \label{tab:human_rubric}
  \small
  \setlength{\tabcolsep}{5pt}
  \renewcommand{\arraystretch}{1.12}
  \begin{tabular}{p{0.24\linewidth}p{0.68\linewidth}}
  \toprule
  \textbf{Criterion} & \textbf{Description} \\
  \midrule
  \multicolumn{2}{c}{\textit{Task 1: Navigation Instructions}} \\
  \midrule
  Correctness
  & How well the instructions match the true trajectory in destination, direction, route constraints, and waypoints. \\
  No hallucination
  & Whether the instructions avoid inventing roads, POIs, landmarks, or environmental details not supported by the trajectory data. \\
  Persona fidelity
  & How naturally and consistently the instructions reflect the assigned persona across all three variants. \\
  Style distinctness
  & How clearly the literal, concise, and chatty variants differ in style rather than only in length. \\
  \midrule
  \multicolumn{2}{c}{\textit{Task 2: Retrieval Queries}} \\
  \midrule
  Retrieval specificity
  & How well the queries capture the distinctive features needed to retrieve this trajectory from a database. \\
  Accuracy
  & How accurately the queries describe the true origin, destination, route properties, timing, and constraints. \\
  No hallucination
  & Whether the queries avoid inventing non-existent places, route constraints, or semantic cues. \\
  \midrule
  \multicolumn{2}{c}{\textit{Task 3: Trajectory Caption}} \\
  \midrule
  Comprehensiveness
  & How completely the caption covers the start, major traversed regions/phases, and endpoint of the trajectory. \\
  Accuracy
  & How faithfully the caption reflects the true route order, direction changes, and spatial progression. \\
  Objectivity / purity
  & Whether the caption stays fully objective, without subjective intent, anthropomorphic framing, or fabricated details. \\
  \bottomrule
  \end{tabular}
\end{table}

LLM and human annotators evaluate the generated language outputs against the underlying
trajectory evidence. For Task 1, annotators jointly inspect the literal, concise,
and chatty navigation instructions. For Task 2, they jointly inspect the three
retrieval queries. For Task 3, they inspect the trajectory caption. Each criterion
is rated on a 1--5 Likert scale, where 5 indicates excellent satisfaction of the
criterion and 1 indicates failure.
Human annotators additionally receive an interactive map visualization of each
trajectory, allowing them to zoom, pan, and inspect the route geometry and
surrounding POIs before scoring.

\begin{table}[t]
    \centering
    \caption{Mean quality scores (1--5 scale) from three LLM judges
    ($N\!=\!2{,}000$ trajectories) and three human judges
    ($N\!=\!100$ Porto trajectories).}
    \label{tab:human_mean_score}
    \small
    \setlength{\tabcolsep}{5pt}
    \begin{tabular}{lcccccc}
    \toprule
    & \multicolumn{3}{c}{\textbf{LLM} ($N\!=\!2\text{K}$)}
    & \multicolumn{3}{c}{\textbf{Human} ($N\!=\!100$)} \\
    \cmidrule(lr){2-4}\cmidrule(lr){5-7}
    \textbf{Criterion} & \textbf{G} & \textbf{P} & \textbf{Q} & \textbf{H1} & \textbf{H2} & \textbf{H3} \\
    \midrule
    \multicolumn{7}{l}{\textit{Task 1: Instruction Generation}} \\
    \quad Correctness & 4.31 & 4.74 & 4.06 & 4.05 & 3.98 & 4.54 \\
    \quad No Hallucination & 4.86 & 4.84 & 4.19 & 3.88 & 3.84 & 4.77 \\
    \quad Persona Fidelity & 4.64 & 4.99 & 4.83 & 4.59 & 4.71 & 4.91 \\
    \quad Style Distinctness & 4.98 & 4.98 & 4.19 & 5.00 & 4.30 & 4.90 \\
    \midrule
    \multicolumn{7}{l}{\textit{Task 2: Retrieval Query}} \\
    \quad Retrieval Specificity & 4.82 & 4.86 & 4.14 & 4.04 & 4.03 & 4.78 \\
    \quad Accuracy & 4.62 & 4.87 & 4.21 & 3.91 & 3.76 & 4.89 \\
    \quad No Hallucination & 4.80 & 4.88 & 4.47 & 3.90 & 3.59 & 4.87 \\
    \midrule
    \multicolumn{7}{l}{\textit{Task 3: Trajectory Captioning}} \\
    \quad Comprehensiveness & 4.90 & 4.99 & 4.63 & 4.13 & 4.16 & 4.99 \\
    \quad Accuracy & 4.53 & 4.98 & 4.52 & 3.90 & 3.77 & 4.98 \\
    \quad Objectivity Purity & 4.90 & 5.00 & 4.79 & 4.85 & 3.89 & 4.99 \\
    \midrule
    \textbf{Mean} & \textbf{4.74} & \textbf{4.91} & \textbf{4.40} & \textbf{4.22} & \textbf{4.00} & \textbf{4.86} \\
    \bottomrule
    \end{tabular}
\end{table}

Table~\ref{tab:human_mean_score} reports mean quality scores from
three LLM judges and three human judges.
All criteria exceed 4.0 on average, indicating high overall
annotation quality.
Human judges are systematically stricter than LLMs, particularly
on \emph{No Hallucination} (human mean $\approx$3.8--4.8 vs.\
LLM $\geq$4.2), suggesting that humans detect subtle spatial
fabrications that LLM judges overlook.
Persona Fidelity and Style Distinctness receive the highest human
scores ($\geq$4.3), confirming that the style and persona guidance
in the generation pipeline produces perceptually distinct outputs.

\begin{table}[t]
    \centering
    \caption{$\pm$1 agreement: LLM pairs on $N\!=\!2{,}000$;
    human pairs and human--LLM cross-agreement on $N\!=\!100$.
    \textbf{H$\leftrightarrow$L} averages all nine human--LLM pairs.}
    \label{tab:human_llm_agreement}
    \small
    \setlength{\tabcolsep}{4pt}
    \begin{tabular}{lccccccc}
    \toprule
    & \multicolumn{3}{c}{\textbf{LLM$\leftrightarrow$LLM} ($N\!=\!2\text{K}$)}
    & \multicolumn{3}{c}{\textbf{H$\leftrightarrow$H} ($N\!=\!100$)}
    & ($N\!=\!100$) \\
    \cmidrule(lr){2-4}\cmidrule(lr){5-7}\cmidrule(lr){8-8}
    \textbf{Criterion} & G$\leftrightarrow$P & G$\leftrightarrow$Q & P$\leftrightarrow$Q & H1$\leftrightarrow$H2 & H1$\leftrightarrow$H3 & H2$\leftrightarrow$H3 & H$\leftrightarrow$L \\
    \midrule
    \multicolumn{8}{l}{\textit{Task 1: Instruction Generation}} \\
    \quad Correctness & 81.7\% & 87.6\% & 99.3\% & 99.0\% & 96.0\% & 89.0\% & 92.1\% \\
    \quad No Hallucination & 94.8\% & 96.1\% & 98.8\% & 98.0\% & 90.0\% & 86.0\% & 93.0\% \\
    \quad Persona Fidelity & 89.1\% & 91.2\% & 99.7\% & 100.0\% & 100.0\% & 99.0\% & 99.1\% \\
    \quad Style Distinctness & 99.6\% & 99.6\% & 99.8\% & 95.0\% & 99.0\% & 94.0\% & 98.7\% \\
    \midrule
    \multicolumn{8}{l}{\textit{Task 2: Retrieval Query}} \\
    \quad Retrieval Specificity & 98.2\% & 99.4\% & 99.7\% & 98.0\% & 94.0\% & 85.0\% & 93.9\% \\
    \quad Accuracy & 89.5\% & 96.1\% & 99.7\% & 98.0\% & 88.0\% & 77.0\% & 90.4\% \\
    \quad No Hallucination & 94.1\% & 94.3\% & 98.8\% & 96.0\% & 91.0\% & 69.0\% & 86.4\% \\
    \midrule
    \multicolumn{8}{l}{\textit{Task 3: Trajectory Captioning}} \\
    \quad Comprehensiveness & 98.8\% & 99.5\% & 100.0\% & 99.0\% & 90.9\% & 84.8\% & 93.6\% \\
    \quad Accuracy & 88.4\% & 93.4\% & 99.9\% & 100.0\% & 84.8\% & 74.7\% & 87.0\% \\
    \quad Objectivity Purity & 97.8\% & 98.4\% & 99.8\% & 91.0\% & 100.0\% & 85.9\% & 94.4\% \\
    \midrule
    \textbf{Mean} & \textbf{93.2\%} & \textbf{95.6\%} & \textbf{99.5\%} & \textbf{97.4\%} & \textbf{93.4\%} & \textbf{84.4\%} & \textbf{92.9\%} \\
    \bottomrule
    \end{tabular}
\end{table}

Table~\ref{tab:human_llm_agreement} measures pairwise $\pm$1
agreement.
LLM--LLM agreement is consistently high (mean 93--99\%), with
P$\leftrightarrow$Q reaching 99.5\%.
Human--human agreement averages 91.7\% but drops on criteria
requiring fine-grained spatial verification (e.g.,
Accuracy H2$\leftrightarrow$H3: 74.7\% for Task~3), reflecting
inherent subjectivity in judging factual completeness.
Cross-modal human--LLM agreement averages 92.9\%, validating that
LLM judges can serve as reliable proxies for scalable quality
control while human review remains essential for detecting
hard-to-verify hallucinations.

\subsubsection{LLM Data Generation Prompt}\label{sec:llm_prompt}

Below we show the system prompt and user prompt template used for
multi-task annotation generation (Claude Sonnet 4.6 as an example;
Gemini and GPT-OSS use equivalent prompts).
Section headers and representative rules are preserved;
exhaustive examples and enumerations are abbreviated with \texttt{[...]}.

\clearpage
\paragraph{System prompt (abbreviated).}
\noindent
\begin{lstlisting}[basicstyle=\ttfamily\scriptsize,
  frame=single, breaklines=true, columns=fullflexible,
  showstringspaces=false]
## 1. Core Directive
Reverse-engineer the user's travel intent from a real driving
trajectory (compressed JSONL with semantics, topology, headings).

* Data Grounding [ZERO TOLERANCE]: ONLY use features in the data.
  Do NOT invent traffic lights, bridges, tunnels, etc.
* Anti-Tour Guide: Focus on end-goal + 1-2 constraints.
  Selectively IGNORE intermediate phases.
* Phase Alignment [ZERO TOLERANCE]: Origin from Phase 0 only;
  destination from last phase only; waypoints from middle only.
* Parrot Ban [ZERO TOLERANCE]: NEVER copy GIS/GNN labels.
  WATERFRONT -> "by the river"; GREEN/PARK -> "the park";
  URBAN/INLAND -> "downtown"; Commercial -> "the shops"; etc.
* No em-dashes or semicolons in any output.
[... persona rules, grammar, narrative handling ...]

## 2. Intent Taxonomy (4 dimensions, 10 scenarios)
  Dim 1 Destination: 1.1 Exact Anchor, 1.2 Fuzzy Semantic
  Dim 2 Waypoint: 2.1 Strict Sequential, 2.2 Flexible, 2.3 Zone
  Dim 3 Route Pref: 3.1 Semantic, 3.2 Topological, 3.3 Orthogonal
  Dim 4 Temporal: 4.1 Time-of-Day, 4.2 Pace/Duration

## 3. Diversity Strategy
* Single intent (minority): Dimension 1 only.
* Composite (majority): Dim 1 once + scenarios from Dims 2-4.
* Anti-Cliche: rotate waypoints (gas, ATM, pharmacy, ...).
[... diversity pools, rotation lists ...]

## 4. Few-Shot Examples
* [BAD] "Navigate to the waterfront area." (parroting labels)
* [GOOD] "Hospital. Now." (ultra-short)
* [GOOD] "No rush. Take me by the river, pull over somewhere
  I can stretch my legs." (leisurely + waypoint)
[... 15+ additional examples ...]

## 5. Three Instruction Styles
  Literal: faithful, explicit.  Concise: ultra-short.
  Chatty: conversational. Each with a DIFFERENT opener.

## 6. Retrieval Queries & Caption
* 3 retrieval queries (Task 2): search-style, NOT navigation.
  Together cover all 4 dimensions.
* 1 caption (Task 3): 3rd-person, factual, present-tense.
[... specificity rules, examples ...]

## 7. Output Format (JSON, no chain-of-thought)
{ "_intent_planning": "...", "_retrieval_planning": "...",
  "instruction_literal": "...", "instruction_concise": "...",
  "instruction_chatty": "...", "retrieval_query_1": "...",
  "retrieval_query_2": "...", "retrieval_query_3": "...",
  "trajectory_caption": "..." }
\end{lstlisting}

\paragraph{User prompt template (per trajectory).}\leavevmode\\
\noindent\begin{minipage}{\linewidth}
\begin{lstlisting}[basicstyle=\ttfamily\scriptsize,
  frame=single, breaklines=true, columns=fullflexible,
  showstringspaces=false, escapeinside={(*}{*)}]
For this trajectory, generate three instructions (literal,
concise, chatty) reflecting these intent type(s):

  (*$\langle$\textit{\textbf{scenario\_labels}}$\rangle$*)  (*\textnormal{\scriptsize e.g.\ ``1.1 + 2.2 + 3.2''}*)

INTENT PLANNING (required): output "_intent_planning" first.

RETRIEVAL ASSIGNMENT (3 queries cover ALL 4 DIMENSIONS):
  - retrieval_query_1 -> (*$\langle$\textit{\textbf{assigned\_dims}}$\rangle$*)
  - retrieval_query_2 -> (*$\langle$\textit{\textbf{assigned\_dims}}$\rangle$*)
  - retrieval_query_3 -> (*$\langle$\textit{\textbf{assigned\_dims}}$\rangle$*)

SPEAKER PERSONA: (*$\langle$\textit{\textbf{persona}}$\rangle$*)  (*\textnormal{\scriptsize e.g.\ ``Impatient and rushed''}*)

STYLE GUIDANCE:
  - literal:  (*$\langle$\textit{\textbf{style}}$\rangle$*)  [length: (*$\langle$\textit{\textbf{len}}$\rangle$*)]
  - concise:  (*$\langle$\textit{\textbf{style}}$\rangle$*)  [length: (*$\langle$\textit{\textbf{len}}$\rangle$*)]
  - chatty:   (*$\langle$\textit{\textbf{style}}$\rangle$*)  [length: (*$\langle$\textit{\textbf{len}}$\rangle$*)]

(*$\langle$\textit{\textbf{optional constraints}}$\rangle$*)  (*\textnormal{\scriptsize waypoint, street name, time context}*)

NARRATIVE MODE -- IGNORE: Do not use text after "Narrative:"
in phase descriptions for facts or phrasing.

Output ONLY a JSON object. No reasoning.

(*$\langle$\textit{\textbf{compressed\_trajectory\_JSON}}$\rangle$*)
\end{lstlisting}
\end{minipage}

\subsection{Qualitative Examples}\label{sec:qual-examples}

\begin{figure*}[htbp]
  \begin{minipage}{0.38\textwidth}
      \centering
      \includegraphics[width=\linewidth]{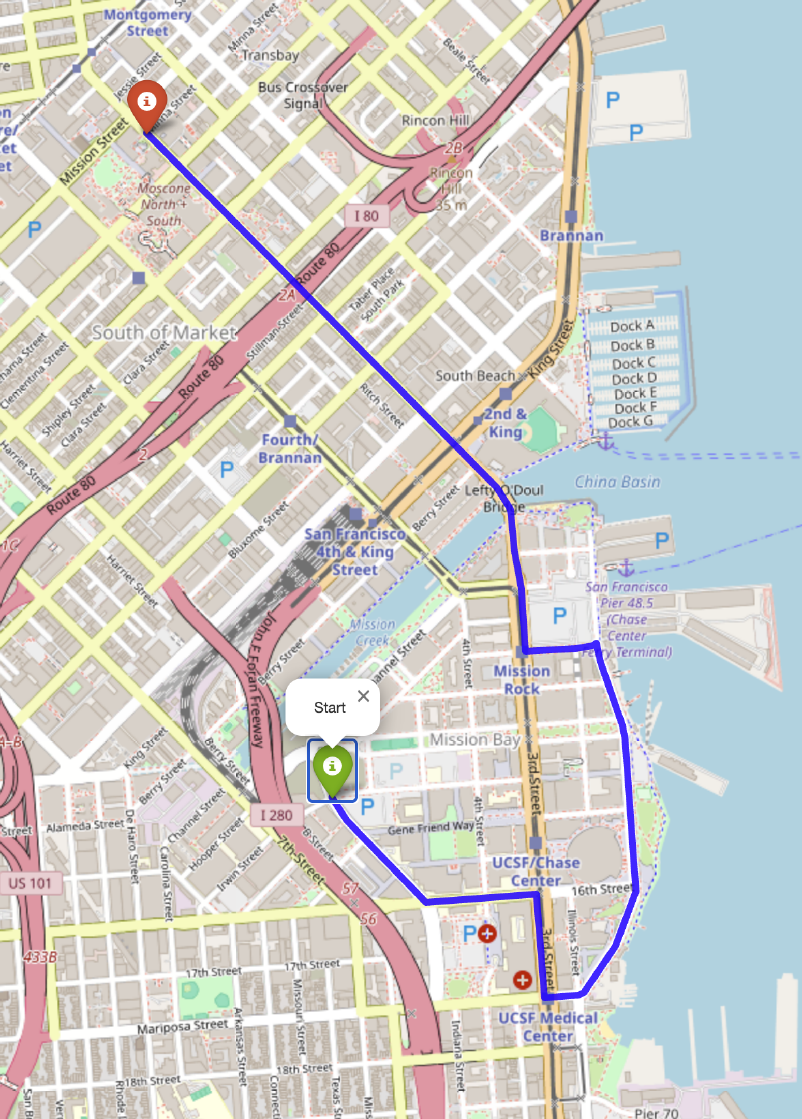}
  \end{minipage}
  \hfill
  \begin{minipage}{0.59\textwidth}
      \small
      \textbf{Task 1 --- Navigation Instructions}
      \;\textcolor{teal}{\scriptsize\textsf{[1.1 Exact Anchor $\cdot$ 2.1 Strict Sequential]}} \\[2pt]
      \textit{Literal:} ``Head to W Hotels via Owens Street, then 16th Street, Terry A. François Boulevard, Mission Rock Street and finish on 3rd Street.'' \\[2pt]
      \textit{Concise:} ``Via 16th, Terry A. François Boulevard, Mission Rock, 3rd to W Hotels.'' \\[2pt]
      \textit{Chatty:} ``Hey, how about take me from Owens Street down 16th Street, swing onto Terry A. François Boulevard, cut through Mission Rock Street and finish on 3rd Street at the W Hotels? It's a quick morning run, so get me there by 10:10.'' \\[4pt]
      \textbf{Task 2 --- Retrieval Queries:} \\[2pt]
      \textit{Query 1:} ``Trips from Owens Street to a beachside spot that pass sequentially through 16th Street, Terry A. François Boulevard and Mission Rock Street.''
      \;\textcolor{purple}{\scriptsize\textsf{[2.1 Strict Sequential $\cdot$ 1.2 Fuzzy Semantic]}} \\[2pt]
      \textit{Query 2:} ``Recorded drives from Owens Street to W Hotels that mainly follow major roads and keep a westward bearing.''
      \;\textcolor{purple}{\scriptsize\textsf{[3.2 Topological/Direct.]}} \\[2pt]
      \textit{Query 3:} ``Morning weekday routes starting at Owens Street around 10 AM and ending at W Hotels.''
      \;\textcolor{purple}{\scriptsize\textsf{[4.1 Time-of-Day]}} \\[4pt]
      \textbf{Task 3 --- Trajectory Caption:} \\[2pt]
      \textit{Caption:} ``Around 10:03 AM on a Monday, the drive departs from Owens Street, heads southwest onto 16th Street, then turns onto Terry A. François Boulevard, continues westward through Mission Rock Street and merges onto 3rd Street, arriving at the W Hotels after roughly seven minutes.''
  \end{minipage}
  \caption{Qualitative example from San Francisco (traj\_id: 475).
  Left: ground-truth trajectory on the road network.
  Right: seven generated annotations spanning three tasks.}
  \label{fig:qual-example-sf}
  \end{figure*}

\begin{figure*}[htbp]
  \begin{minipage}{0.38\textwidth}
      \centering
      \includegraphics[width=\linewidth]{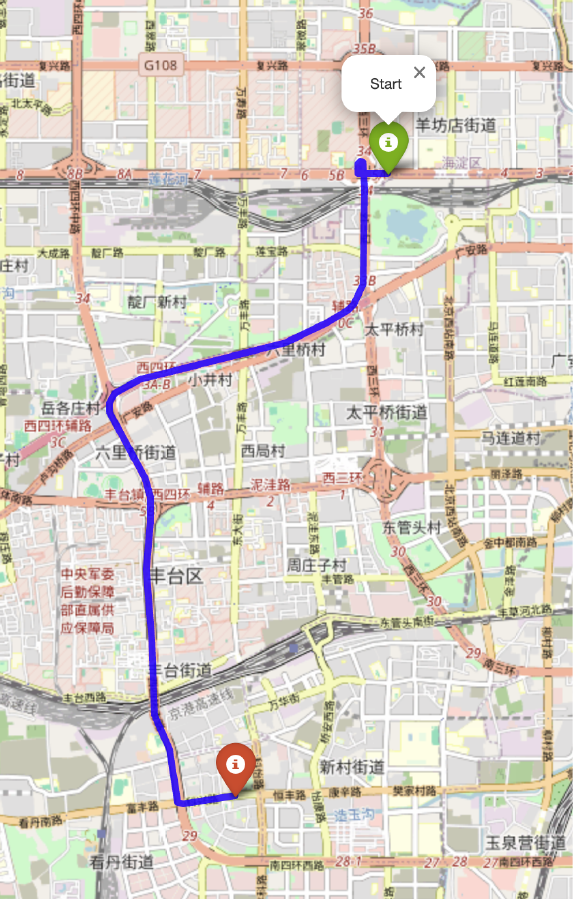}
  \end{minipage}
  \hfill
  \begin{minipage}{0.59\textwidth}
      \small
      \textbf{Task 1 --- Navigation Instructions}
      \;\textcolor{teal}{\scriptsize\textsf{[1.1 Exact Anchor]}} \\[2pt]
      \textit{Literal:} ``Head to Yonghe King near the KFC.'' \\[2pt]
      \textit{Concise:} ``Yonghe King. Quick.'' \\[2pt]
      \textit{Chatty:} ``Man, I need to get to Yonghe King. Just take the usual route past the KFC.'' \\[4pt]
      \textbf{Task 2 --- Retrieval Queries:} \\[2pt]
      \textit{Query 1:} ``Recorded Sunday night trips from Lianhua Bridge station to Yonghe King that were taken after 10 PM.''
      \;\textcolor{purple}{\scriptsize\textsf{[4.1 Time-of-Day]}} \\[2pt]
      \textit{Query 2:} ``Historical routes from Lianhua Bridge station to Yonghe King.''
      \;\textcolor{purple}{\scriptsize\textsf{[1.1 Exact Anchor]}} \\[2pt]
      \textit{Query 3:} ``Trips from Lianhua Bridge station to Yonghe King that travel via Xisihuan South Road and then continue onto Kexing Road, heading east on the ring road and turning north onto Kexing Road.''
      \;\textcolor{purple}{\scriptsize\textsf{[2.1 Strict Sequential $\cdot$ 3.2 Topological/Direct.]}} \\[4pt]
      \textbf{Task 3 --- Trajectory Caption:} \\[2pt]
      \textit{Caption:} ``Starting at Lianhua Bridge station in the western suburbs, the 13-minute drive first heads west on local streets, then swings east onto a major ring road and later onto its middle segment. It continues eastward along the southern part of that ring road before turning north onto the street with the fast-food outlet, where it finishes at the Yonghe King fast-food outlet next to a KFC. The route weaves through mixed residential blocks, passes a few parks, and stays on the main thoroughfares throughout.''
  \end{minipage}
  \caption{Qualitative example from Beijing (traj\_id: 9331).
  Left: ground-truth trajectory on the road network.
  Right: seven generated annotations spanning three tasks.}
  \label{fig:qual-example-bj}
\end{figure*}

\begin{figure*}[htbp]
  \begin{minipage}{0.38\textwidth}
      \centering
      \includegraphics[width=\linewidth]{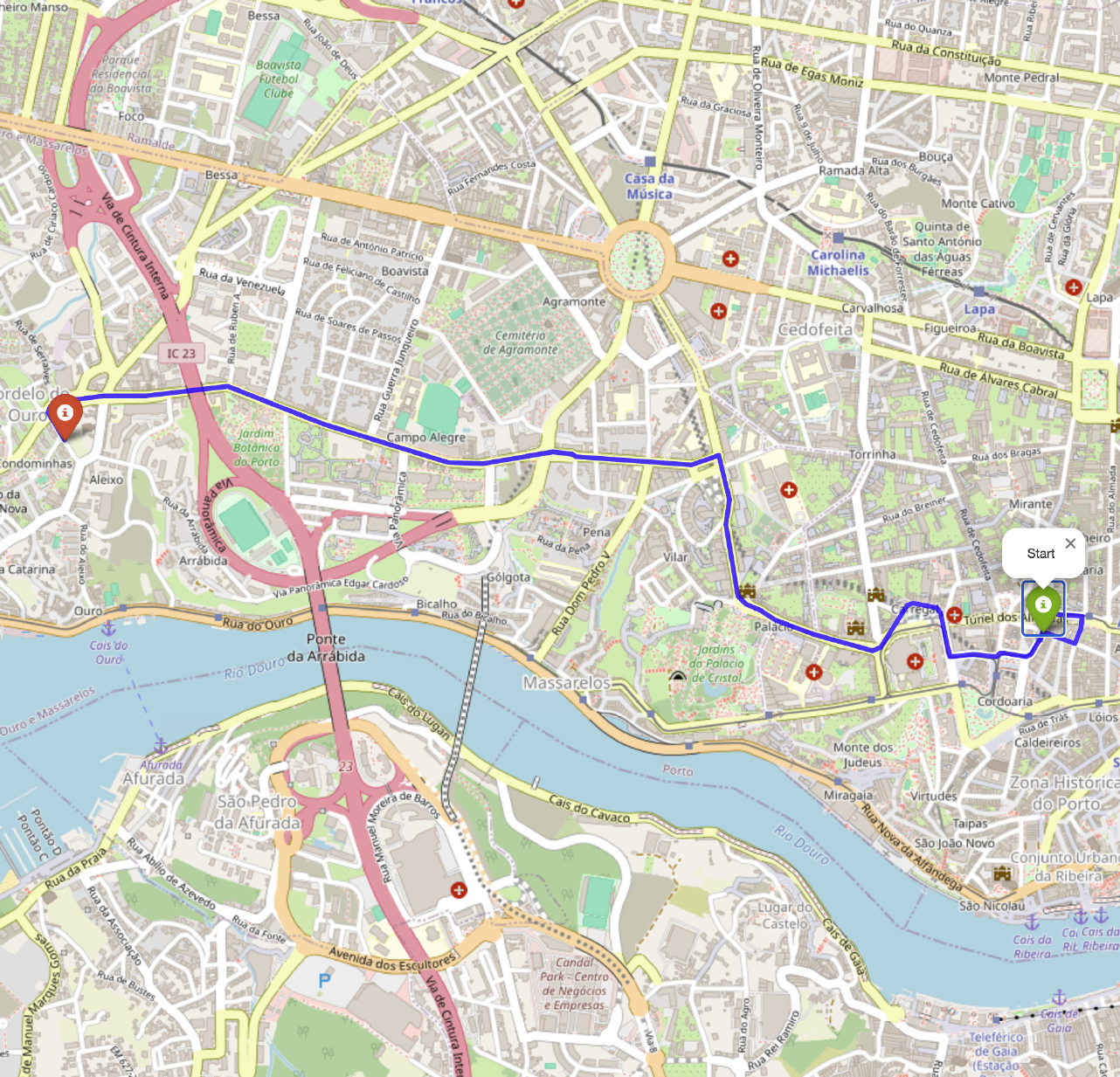}
  \end{minipage}
  \hfill
  \begin{minipage}{0.59\textwidth}
      \small
      \textbf{Task 1 --- Navigation Instructions}
      \;\textcolor{teal}{\scriptsize\textsf{[1.1 Exact Anchor $\cdot$ 3.3 Orthogonal Comp. $\cdot$ 4.1 Time-of-Day $\cdot$ 4.2 Pace/Duration]}} \\[2pt]
      \textit{Literal:} ``To Ponte D via the park-side roads. It's 2 AM, traffic's dead, so just push straight through. No detours, no stops.'' \\[2pt]
      \textit{Concise:} ``Ponte D. Night run. Park roads. Fast.'' \\[2pt]
      \textit{Chatty:} ``Okay so it's the middle of the night, just get me to Ponte D. Cut through that green stretch, it's dead quiet at this hour anyway. Don't mess around with the busy streets back in Cedofeita, just punch through and get there. Nine minutes, tops.'' \\[4pt]
      \textbf{Task 2 --- Retrieval Queries:} \\[2pt]
      \textit{Query 1:} ``Trips from the Cedofeita area to Ponte D in Lordelo do Ouro e Massarelos that pass continuously through the park and tree-lined corridor without any recorded stops along the way.''
      \;\textcolor{purple}{\scriptsize\textsf{[2.3 Pass-through Zone]}} \\[2pt]
      \textit{Query 2:} ``Recorded routes from Cedofeita to the Ponte D area in Lordelo do Ouro e Massarelos that favor the quieter, greener by the riverside and park-side streets rather than the denser commercial or high-traffic roads.''
      \;\textcolor{purple}{\scriptsize\textsf{[3.1 Semantic Constraints]}} \\[2pt]
      \textit{Query 3:} ``Historical drives ending at Ponte D in Lordelo do Ouro e Massarelos, originating from the Cedofeita district, recorded in the early-morning hours around 2 AM on a weekend night when roads are largely empty.''
      \;\textcolor{purple}{\scriptsize\textsf{[1.1 Exact Anchor $\cdot$ 4.1 Time-of-Day]}}
  \end{minipage}

  \vspace{4pt}
  \noindent\small
  \textbf{Task 3 --- Trajectory Caption:}\;
  \textit{Caption:} ``Starting in the Cedofeita district around 2:05 AM on a Saturday night, the route heads southeast before pivoting northwest and west, passing through a mix of dense urban streets near the city center and transitioning into the docks and park corridor of Lordelo do Ouro e Massarelos. The route travels along Rua do Campo Alegre for several phases, moving through tree-lined and park-adjacent zones including areas near the University of Porto campus, before concluding at the Ponte D area on Rua de Diogo Botelho in Lordelo do Ouro e Massarelos. The full trip covers 12 phases in approximately 9 minutes, with the bulk of travel time spent in the quieter green and the riverside zones during late-night, low-traffic conditions.''

  \caption{Qualitative example from Porto (traj\_id: 1373490).
  Top-left: ground-truth trajectory. Top-right: Task 1 and 2 annotations. Bottom: Task 3 caption.}
  \label{fig:qual-example-porto}
\end{figure*}

\newpage

\subsection{Limitations}\label{sec:limitations}

TrajPrism currently covers three cities with English-language
annotations (except for native-language place names). The reproducible RIR pipeline facilitates extension
to additional cities and languages.
Travel-intent instructions are LLM-synthesized with multi-stage
quality control, which may introduce distributional biases
compared to organically collected user queries.
All trajectories are map-matched GPS traces. 
Indoor, pedestrian, and multimodal transportation mode settings are not yet addressed.


\newpage

\end{document}